\documentclass[12pt]{article}
\usepackage{graphicx}
\usepackage{latexsym}
\usepackage{color}
\usepackage{url}

\newcommand{\be}{\begin{equation}}
\newcommand{\bea}{\begin{eqnarray}}
\newcommand{\ee}{\end{equation}}
\newcommand{\eea}{\end{eqnarray}}
\def \blank{\mbox{}}

\def\r{\mbox{{\bf x}}}

\def\R{\mbox{{\bf R}}}

\def\DF{\mbox{{\bf DF}}}

\def\X{\mbox{{\bf X}}}
\def\x{\mbox{{\bf x}}}

\def\Y{\mbox{$\bf{Y}$}}
\def\y{\mbox{$\bf{y}$}}
\def\v{\mbox{$\bf{v}$}}

\def\p{\mbox{$\bf{p}$}}

\def\f{\mbox{$\bf{f}$}}
\def\F{\mbox{$\bf{F}$}}

\def\r{\mbox{$\bf{r}$}}

\graphicspath{%
    {converted_graphics/}% inserted by PCTeX
    {/}% inserted by PCTeX
    {E:/machines_da/lor96d11data/}% inserted by PCTeX
    {H:/machines_da/lor96d11data/}% inserted by PCTeX
    {E:/machines_da/machine_pr_data/}% inserted by PCTeX
    {H:/machines_da/machine_pr_data/}% inserted by PCTeX
    {H:/machines_da/machine_pr_data/data_060417/}% inserted by PCTeX
    {E:/machines_da/machine_pr_data/data_061617/}% inserted by PCTeX
    {E:/machines_da/machine_pr_data/data_061917/}% inserted by PCTeX
    {E:/machines_da/machine_pr_data/data_062717/}% inserted by PCTeX
    {E:/machines_da/machine_pr_data/data_062917/}% inserted by PCTeX
}
\begin{document}

\begin{center}

\bigskip
\bigskip
{\Large  {\bf Machine Learning, Deepest Learning: Statistical Data Assimilation Problems}}\\
\bigskip

\bigskip
\bigskip
Henry Abarbanel,\\
Department of Physics\\
and\\
Marine Physical Laboratory (Scripps Institution of Oceanography)\\
Center for Engineered Natural Intelligence\\
habarbanel@ucsd.edu\\
\bigskip
 %University of California, San Diego \\
%9500 Gilman Drive, Mail Code 0374\\
%La Jolla, CA 92093-0374, USA\\
\bigskip
Paul Rozdeba\\
\bigskip
and\\
\bigskip
Sasha Shirman\\
\bigskip
\bigskip
\bigskip
Department of Physics\\
 University of California, San Diego \\
9500 Gilman Drive, Mail Code 0374\\
La Jolla, CA 92093-0374, USA\\
\bigskip 
\bigskip 

{\bf Draft of \today}
\end{center}
\newpage 

%\newpage 
%\tableofcontents 
%\newpage 

\section{Abstract}

We formulate a strong equivalence between machine learning, artificial intelligence methods and the formulation of statistical data assimilation as used widely in physical and biological sciences. The correspondence is that layer number in the artificial network setting is the analog of time in the data assimilation setting.
Within the discussion of this equivalence we show that adding more layers (making the network deeper) is analogous to adding temporal resolution in a data assimilation framework.

How one can find a candidate for the global minimum of the cost functions in the machine learning context using a method from data assimilation is discussed. Calculations on simple models from each side of the equivalence are reported. 

Also discussed is a framework in which the time or layer label is taken to be continuous, providing a differential equation, the Euler-Lagrange equation, which shows that the problem being solved is a two point boundary value problem familiar in the discussion of variational methods. The use of continuous layers is denoted ``deepest learning''. These problems respect a symplectic symmetry in continuous time/layer phase space. Both Lagrangian versions and Hamiltonian versions of these problems are presented. Their well-studied implementation in a discrete time/layer, while respected the symplectic structure, is addressed. The Hamiltonian version provides a direct rationale for back propagation as a solution method for the canonical momentum.

\section{Introduction}
Through the use of enhanced computational capability two, seemingly unrelated, `inverse' problems have flourished over the past decade. One is machine learning in the realm of artificial intelligence~\cite{ai_agi17,goodfellow16,deep15} with developments that often go under the name ``deep learning''. The other is data assimilation in the physical and life sciences. This describes the transfer of information from observations to models of the processes producing those observations~\cite{bennett92,even,abar13}. 

This paper is directed towards demonstrating that these two areas of investigation are the same at a fundamental level. Each is a statistical physics problem where methods utilized in one may prove valuable for the other. The main goal of the paper is to point out that many developments in data assimilation can be utilized in the arena of machine learning. We also suggest that innovative methods from machine learning may be valuable in data assimilation.

Two areas of focus are attended to here: (1) a variational annealing (VA) method for the action (cost function) for machine learning or statistical data assimilation that permits the location of the apparent global minimum of that cost function. (2) The notion of analyzing each problem in continuous time or layer, which we call {\tt deepest learning}, wherein it is clear that one is addressing a two point boundary value problem~\cite{ye2015physrev,gfomin} with an underlying symplectic structure. Methods abound for solving such two point boundary value problems~\cite{press} and for assuring that symplectic structures are respected when time (or layer) is discretized. These may be quite fruitful in machine learning.

This paper primarily discusses multilayer perceptrons or feedforward networks~\cite{goodfellow16} though it also makes it is clear that the discussion carries over to recurrent networks as well~\cite{jordan,elman,parlos}.

\section{Background}
\subsection{Machine Learning; Standard Feedforward Neural Nets}
We begin with a brief characterization of simple architectures in feedforward neural networks~\cite{ai_agi17,goodfellow16,deep15}. The network we describe is composed of an input layer $l_0$ and output layer $l_F$ and hidden layers $l_1, l_2, ..., l_F - 1$. Within each layer we have $N$ active units, called `neurons', each of which has $d$ degrees-of-freedom. (Usually d is chosen to be 1.) 

Each layer has $N$ neurons with $d$ degrees of freedom, so each layer has $D = Nd$ degrees-of-freedom. For our purposes the `neurons' in each layer are the same structure. This can be generalized to different numbers and different types of neurons in each layer at the cost of a notation explosion.

Data is available to layer $l_0$ and layer $l_F$ in M pairs of $L$-dimensional input, at $l_0$, and output, at $l_F$. These are sets of vectors: $\{y_r^{(k)}(l_0), y_r^{(k)}(l_F)\}$ where $k = 1,2, ..., M$ labels the pairs, $r$ is an index on the $L$-dimensional data $r = 1, 2, 3, ..., L \le D$.

The activity of the units in each hidden layer $l$, $x^{(k)}_{\alpha}(l);\;l_0 < l < l_F$, is determined by the activity in the previous layer. The index $\alpha$ combines the neuron number $j$ and the neuron degrees-of-freedom $a$ into one label: $\alpha = 1,2,...,Nd = D$; $j=1,2,...N;\;a=1,2,...d$.  
This connection is described by the nonlinear function $f_{\alpha}(\bullet)$ via
\small
\be 
x^{(k)}_{\alpha}(l) = f_{\alpha}(\x(l-1),l) = f_{\alpha}\biggl(\sum_{\beta=1}^{Nd = D}W_{\alpha \,\beta}(l)x^{(k)}_{\beta}(l-1)\biggr),
\label{networkml}
\ee
\normalsize
where $\x^{(k)}(l)= \{x^{(k)}_{\alpha}(l)\} = \{x^{(k)}_1(l),x^{(k)}_2(l),...,x^{(k)}_{Nd}(l)\}$. The summation over weights $W_{\alpha \, \beta}(l)$ determines how the 
activities in layer $l-1$ are combined before allowing $f_{\alpha}(\bullet)$ to act, yielding the activities at layer $l$.
There are numerous choices for the manner in which the weight functions act as well as numerous choices for the nonlinear functions, and we direct the reader to the references for the discussion of the virtues of those choices~\cite{ai_agi17,goodfellow16,deep15}.

%We denote a collection of layers $\{l_0,l_1,...,l_{N1}\}$ as an `epoch' through which initial values of the activations at the input layer, $x_{\alpha}(l_0)$ are transported through the network until reaching an output layer at which the activities are $x_{\beta}(l_{N1})$. These activities are then compared with data which we call $y_{r}(l_{N1})$. The index $r$ could run from $r = 1,2,...,Nd = D$ in which case there are enough observations at $l_{N1}$ to compare with each of the activities $x_{\alpha}(l_{N1})$. We will allow here for the possibility that there are fewer observations than $D$ at $l_{N1}$, and we allow $r = 1,2,...,L1;\;L1 \le D$.

At the input and the output layers $l_0, l_{F}$ the network activities are compared to the observations, and the network performance is assessed using an error metric, often a least squares criterion, or cost function
\small 
\be 
C_M = \frac{1}{M}\sum_{k=1}^M \biggl \{\frac{1}{2\,L}\sum_{r = 1}^{L} R_m(r) \biggl([x^{(k)}_r(l_0) - y^{(k)}_r(l_0)]^2 + [x^{(k)}_{r}(l_F) - y^{(k)}_r(l_F)]^2\biggr) \biggr \},
\label{cost}
\ee 
\normalsize
where $R_m(r)$ permits an inhomogeneous weighting in the comparison of the network activities and the data. Minimization of this cost function over all $x^{(k)}_{\alpha}(l)$ and weights $W_{\alpha\,\beta}(l)$, {\em subject to} the network model Eq. (\ref{networkml}), is used to determine the weights, the variables $x^{(k)}_{\alpha}(l)$in all layers, and any other parameters appearing in the architecture of the network.

%We could, alternatively, generalize the learning procedure (estimating parameters and unobserved state variables at $l_{N1}$) by adding a sequence of epochs beyond the first. The second epoch has layers $\{l_{N1+1},l_{N1+2},...,l_{N2}\}$. We call the layers at which comparisons are made with data as $\lambda_1 = l_{N1}, \lambda_2 = l_{N2}, ..., \lambda_k = l_{Nk},...,\lambda_F = l_{NF}$. Our F epochs end at the final output layer $l_{NF}$.
%Use the output of epoch $k$ as input to epoch $k+1$. Think of this as repeatedly informing the network where its output should land in state space $\x(l)$ by starting anew at the location it reached before the comparison was made at $l_{Nk} = \lambda_k$.

%Our focus here is on the minimization of $C_K(l_{N1})$ varying all $x_{\alpha}(l)$ and the $W_{\alpha \, \beta}(l)$ using k = 1,2,...,K $\{y^{(k)}_r(l_0), y^{(k)}_r(l_{N1}\}$ data pairs.

Before moving along to data assimilation, we note that one wishes to find the global minimum of the cost function Eq. (\ref{cost}), which is a nonlinear function of the neuron activities, the weights and any other parameters in the functions at each layer. This is an NP-complete problem~\cite{murty87} and as such suggests one cannot find a global minimum of the machine learning problem, as set, unless there is a special circumstance. We will see just such a circumstance in a data assimilation problem equivalent to machine learning tasks.

The machine learning problem as described here assumes there is no error in the model itself, so that the minimization of the cost function Eq. (\ref{cost}) is subject to strong equality constraints through the model. This results in the output at layer $l_F$ $\x^{(k)}(l_F)$ being a very complicated function of the parameters in the model and the activities at layers $\x^{(k)}(l \le l_F)$. This is likely connected with the many local minima associated with the NP-complete nature of the search problem. 

We introduce a variational annealing (VA) method in the next section which regularizes this by moving the equality constraint into the cost function via a penalty function. This introduces a hyperparameter allowing us to systematically vary the complexity of the search process.

\subsection{Standard Statistical Data Assimilation}

Now we describe the formulation of a statistical data assimilation problem. 

In data assimilation observations are made of a sparse set of dynamical variables, associated with a model of the processes producing the observations. This will allow the estimation of parameters in the model and of the unobserved state variables of the model. The goal is to estimate any unknown parameters in the model, and because not all of the dynamical state variables in the model may be observed, to estimate those unmeasured state variables as well. After a certain time window in which information is transferred to the model, we have an estimate of the full model including an initial condition for all state variables, and predictions are made with the completed model and compared to new observations. This validation by prediction is essentially the same as the question of generalization as addressed in machine learning~\cite{goodfellow16}.

%As we proceed, it will become apparent that the translation from the machine learning example above to this example is strikingly simple. 
%Then we will describe how important questions, quite similar in the two tasks, machine learning and data assimilation.

In data assimilation, one has a window in time $[t_0,t_F]$ during which observations are made at times $t = \{\tau_1, \tau_2, ..., \tau_F\}$ which lie in $[t_0 \le \tau_s \le t_F]; s = 1, 2, ..., F$. At each observation time $L$ measurements $y_l(\tau_s);\;l=1,2,...,L$ are made, $L \le D$. Through knowledge of the measurement instruments the observations are related to the state variables of the model via `measurement functions' $h_l(\x):\; y_l(\tau_k) = h_l(\x(\tau_k)); \; l=1,2,...,L$. 

Using what knowledge we have of the processes producing the observations, we develop a dynamical model for the state variables. These satisfy a set of $D$ dynamical differential equations
\be 
\frac{dx_a(t)}{dt} = F_a(\x(t),t);\;a=1,2,...,D.
\label{model}
\ee
The time dependence of the vector field $\F(\x,t)$ may come from external forcing functions driving the dynamics.

This set of differential equations will necessarily be represented in discrete time when solving them numerically, resulting in a map $x_a(t_{k}) \to x_a(t_{k+1}) = f_a(\x(t_k), t_k)$ in which the discrete time vector field $f_a(\bullet)$ is related to $F_a(\x,t)$ via the integration procedure one chooses for Eq. (\ref{model}). 

Starting from an initial condition at $t_0$, $x_a(t_0)$ we use the discrete time model
\be 
x_a(t_{k+1}) = f_a(\x(t_k), t_k)
\label{modeldiscrete}
\ee 
to move forward to the first observation time $\tau_1$, then to $\tau_2$, ... eventually reaching the end of the observation window at $t_F$. Altogether by making N model integration steps in each of the intervals $[\tau_n, \tau_{n+1}]$ we make $(F+1)N$ time steps.: $t_0 \to \tau_1 \to \tau_2 ...\to \tau_F \to t_F$.

The measurements are noisy, and the models have errors; so this is a statistical problem.
Our goal is to construct the conditional probability distribution, $P(\X|\Y)$, of the model states $\X(t_F) = \{\x(t_0), \x(t_1), ..., \x(t_{N}), ..., \x(t_F)\}$,
conditioned on the $LF$ measurements $\Y(\tau_F) = \{\y(\tau_1),\y(\tau_2),..., \y(\tau_k),...,\y(\tau_F)\}$. 

Assuming the transition to the model state at time $t_{k+1}$ depends only on the model state at time $t_k$ (that is, the dynamics in Eq. (\ref{modeldiscrete}) is Markov) and using identities on conditional probabilities~\cite{abar13}, one may write the action $A(\X) = -\log[P(\X|\Y)]$ (suppressing the dependence on the observations $\Y$ in $A(\X)$) as
\bea
&&A(\X) = - \sum_{n=1}^{F} CMI[\X(\tau_n),\y(\tau_n)|\Y(\tau_{n-1})] \nonumber \\
&&- \sum_{n=0}^{N(F+1) -1} \log [P(\x(t_{n+1})|\x(t_n))] - \log[P(\x(t_0))],
\label{genaction}
\eea
where the conditional mutual information is given as~\cite{fano} $CMI(a,b|c) = \log\biggl [\frac{P(a,b|c)}{P(a|c)\,P(b|c)}\biggr].$
If the model is error free, $P(\x(t_{n+1})|\x(t_n))$ is a delta function: $P(\x(t_{n+1})|\x(t_n)) = \delta^D(\x(t_{n+1}) - \f(\x(t_n), t_n))$.

With a representation of $P(\X|\Y)$ we may evaluate conditional expected values of functions $G(\X)$ on the path $\X(NF)$ of the model through the observation window $[t_0,t_F]$ as
\be 
E[G(\X)|\Y] = \left \langle G(\X) \right \rangle  = \frac{\int d\X \,G(\X) \exp[-A(\X)]}{\int d\X \exp[-A(\X)]},
\label{expected}
\ee 
in which
\bea
 &&A(\X) = - \sum_{n=1}^{F} \log[P(\y(\tau_n)|\X(\tau_n),\Y(\tau_{n-1})] \nonumber \\
&&- \sum_{n=0}^{N(F+1) -1} \log P[(\x(t_{n+1})|\x(t_n))] - \log[P(\x(t_0))],
\eea 
and terms depending only on the observations were canceled between numerator and denominator in the expected value.

If the observations at the times $\tau_k$ are independent, and if the measurement function is the identity $y_r(\tau_k) = h_r(\x(\tau_k)) = x_r(\tau_k)$, and if the noise in the measurements is Gaussian, with a diagonal inverse covariance matrix $R_m(r,\tau_k)$, the first term in the action $A(\X)$, the {\tt measurement error term}, takes the form
\be 
\sum_{n=1}^{F} \sum_{r=1}^L \frac{R_m(r, \tau_n)}{2} \biggl(x_r(\tau_n) - y_r(\tau_n)\biggr)^2.
\ee
If no measurement is made at $\tau_k$, $R_m(\r,\tau_k) = 0$.

If the error in the model Eq. (\ref{modeldiscrete}) is taken as additive and Gaussian with diagonal inverse covariance matrix $R_f(a)$ the second term in $A(\X)$, the {\tt model error term}, becomes
\be 
 \sum_{n=0}^{N(F+1)-1} \sum_{a=1}^{D} \frac{R_f(a)}{2} \biggl ( x_a(t_{n+1}) - f_a(\x(t_n), t_n)\biggr)^2.
\ee 

In each term constants having to do with normalizations of the Gaussians cancel in the expected value. If we keep these constants and take the limit $R_f(a) \to \infty$, we would restore the delta function in the dynamics. 

Finally, if one accepts ignorance of the distribution of initial conditions $P(\x(t_0))$ and selects it as uniform over the dynamical range of the model variables, $\left \langle G(\X) \right \rangle$ is evaluated with
\small
\bea 
&&A_0(\X) = \sum_{n=1}^{F} \sum_{r=1}^L \frac{R_m(r, \tau_n)}{2} \biggl(x_r(\tau_n) - y_r(\tau_n)\biggr)^2 \nonumber \\
&&+ \sum_{n=0}^{N(F+1)-1} \sum_{a=1}^{D}  \frac{R_f(a)}{2} \biggl ( x_a(t_{n+1}) - f_a(\x(t_n),t_n)\biggr)^2.
\label{action0}
\eea

\be 
\left \langle G(\X) \right \rangle = \frac{\int d\X \, G(\X) \exp[-A_0(\X)]}{\int d\X \exp[-A_0(\X)]},
\label{expected0}
\ee 
\normalsize 
This is the desired connection between the machine learning formulation (with model error) and the statistical data assimilation formulation: identify layer labels as time $l \Leftrightarrow t$. 

This alone could be of passing interest. However, there is much in the connection that may be of some utility. We will discuss these items in the data assimilation language, but the translation should be easy at this point. For statistical data assimilation we call this $A_0(\X)$ the standard model.

The critical suggestion here relative to standard practice in machine learning~\cite{ai_agi17,goodfellow16,deep15}, is that by allowing $R_f$ to be finite from the outset--so acknowledging model error--we may add an additional tool for exploration
in machine learning environments where typically no account for model error is introduced. Further, the hyper-parameter $R_f$ serves as a regulating device for the complexity of the surface in path space in which the estimation of states and parameters occurs. 

\section{Data Assimilation Developments of Use in Machine Learning}

\subsection{Finding the Global Minimum}

The key to establishing estimates for unobserved state variables ($L < D$) and for unknown parameters in the model is to perform, approximately of course, the integral Eq. (\ref{expected0}). One can do this using Monte Carlo methods~\cite{landau} or by the method of Laplace~\cite{laplace,laplace2}. In the Laplace method one seeks minima of the action $A_0(\X)$, Eq. (\ref{action0}). The integral is not Gaussian.  If it were, we would just do it. As the functions $f_a(\bullet)$ are nonlinear, we must perform a numerical evaluation of Eq. (\ref{expected}).

The Laplace method approximates the integral with contributions from the lowest minima of the action, if one can find them. Minima associated with paths $\X$ having larger action give exponentially smaller contributions to expected values, Eq. (\ref{expected0}), than paths with smaller action. This allows one to circumvent a search for the global minimum if the parameters, hyperparameters~\cite{goodfellow16}, and other aspects of the model and data yield a set of action levels connected with minima of the action such that one path yields an action level much smaller than any other path. For numerical accuracy one may use that smallest minimum path (comprised of parameters and `hidden' (unobserved) state variables) neglecting larger minima of the action.

We have developed a variational annealing (VA) approach~\cite{ye2014precision,ye2015physrev}, to finding the path with the smallest value of the action. While we have no proof that the global minimum is found, our numerical results indicate this may be the case. The VA method produces a set of minima of the action giving a numerical clue as to the roughness of the surface in path $\X$ space. 

In data assimilation the surface depends, among other items, on the number of measurements $L$ at each observation time $\tau_k$, on the hyper-parameter $R_f$, and on the number of model time steps between measurement times $\tau_n$. This translates directly to the analogous machine learning problem with time $\to$ layer. As the number of model time steps between measurement times increases, the number of hidden layers increases and the model architecture deepens.

VA proceeds by a kind of numerical continuation~\cite{allgower} in $R_f$ of the requirement that varying over all $\X$ and all parameters in $A_0(\X)$ minimizes $A_0(\X)$. The procedure begins by taking $R_f \to 0$, namely the complete opposite of the value found in usual machine learning where $R_f \to \infty$ (deterministic, error free layer to layer maps) from the outset. In the $R_f = 0$ limit, the action is just a quadratic function of the model variables $\x(\tau_k)$ at the times measurements are made, and the minimization is simple: $x_r(\tau_k) = y_r(\tau_k)$ for the $r = 1,2,..., L \le D$ data presented at the input and output layers. The minimum can be degenerate as we know only $L \le D$ values for the state variables. 

At the first step of VA we choose as a solution to the optimization problem $x_r(\tau_k) = y_r(\tau_k)$ and select the other $D - L$ states as drawn from a uniform distribution with ranges known from the dynamical range of the state variables. One can learn that well enough by solving the underlying model forward for various initial conditions. We make this draw K times, and now have K paths $\X^0$ as candidates for the VA procedure.

Now we select a small value for $R_f$, call it $R_{f0}$. With the previous K paths $\X^0$ as K initial choices in our minimization algorithm, we find K paths $\X^1$ for the minimization problem with $R_f = R_{f0}$. This gives us K values of the action $A_0(\X^1)$ associated with the new paths $\X^1$.

Next we increase the value of $R_f$ to $R_f = R_{f0}\alpha$ where $\alpha > 1$. (We have found values of $\alpha$ in the range 1.1 to 2 to be good choices). For this new value of $R_f$, we perform the minimization of the action starting with the K initial paths $\X^1$ from the previous step to arrive at K new paths $\X^2$. Evaluating the action on these paths $A_0(\X^2)$ now gives us an ordered set of actions that are no longer as degenerate. Many of the paths $\X^2$ may give the same numerical value of the action, however, typically the `degeneracy' lies within the noise level of the data $\approx (1/\sqrt{R_m})$. 

This procedure is continued until $R_f$ is `large enough' which is indicated by at least one of the action levels becoming substantially independent of $R_f$. 
%The action levels associated with other paths typically continue to grow with $R_f$. 
As a check on the calculation, we observe that if the action $A_0(\X)$ is independent of $R_f$, its expected value is that of the measurement error term. As the measurement errors were taken to be Gaussian, this term in the action is distributed as $\chi^2$, and its expected value is readily evaluated. If the action levels are at this expected value of $\chi^2$ for large $R_f$, the procedure is consistent and no further increases in $R_f$ are required.

Effectively VA starts with a problem ($R_f = 0$) where the global minimum is apparent and systematically tracks it and many other paths through increases in $R_f$. In doing the `tracking' of the global minimum, one must check that the selected value of $\alpha$ is not too large lest one leave the global minimum and land in another minimum. Checking the result using smaller $\alpha$ is worthwhile.

It is important to note that simply starting with a large value of $R_f$ places one in the undesirable situation of the action $A_0(\X)$ having multiple local minima into which any optimization procedure is quite likely to fall.

In the dynamical problems we have examined, one typically finds that as the number of measurements $L$ at each $\tau_k$ is increased, fewer and fewer minima of the action remain and when $L$ is large enough there is one minimum. This we attribute to the additional information from the augmented set of measurements, and this will be manifest in the discussion below where the additional information effectively controls unstable directions in the phase space.

\subsection{Smallest Minimum; Not Necessarily a Convex Action}

As our goal is to provide accurate estimations of the conditional expected value of functions $G(\X)$ where $\X$, a path in model space, is distributed  as $\exp[-A(\X)]$, we actually do not require convexity of $A(\X)$ as a function in path space. From the point of view of accurately estimating expected values, it is sufficient that the lowest action level be {\bf much} smaller than the second lowest action level. If the action value at the lowest level $A(\X_{\mbox{lowest}})$ is much smaller than the action value at the next minimum $A(\X_{\mbox{second lowest}})$, then by a factor $\exp[-\{A(\X_{\mbox{lowest}}) - A(\X_{\mbox{second lowest}})\}]$, the lowest path $\X_{\mbox{lowest}}$ dominates the integral to be done and provides a sensible choice for the path at which to evaluate the integral. 

\section{Examples from Feedforward Neural Networks and from Data Assimilation}

In this section we examine one example from multi-layer perceptrons and one example from statistical data assimilation. The latter utilizes a differential equation model introduced by Lorenz in 1996~\cite{lor96} which permits one to easily increase the number of dimensions of the phase space, to easily select the number of observations within a designated measurement window, and to easily choose the number of model evaluations between measurement times. The latter is analogous to increasing the number of layers in a multi-layer perceptron. 

In each case we perform a `twin experiment'. We use a model to generate solutions from some initial conditions. These solutions, when Gaussian noise is added to them, become our noisy data. Using the noisy data we use VA to estimate the unobserved state variables (hidden layer variables) and parameters/weights. 

\subsection{Data Assimilation for Lorenz96 Model}

We begin by examining the dynamical equations introduced by~\cite{lor96}:
\be 
\frac{dx_a(t)}{dt} = x_{a-1}(t)(x_{a+1}(t) - x_{a-2}(t)) - x_a(t) + \nu
\label{lorenz96}
\ee
and $a=1,2,...,D$; $x_{-1}(t) = x_{D-1}(t)$; $x_0(t) = x_D(t)$; $x_{D+1}(t) = x_1(t)$. $\nu$ is a fixed parameter which we take to be 10.0 where the solutions to the dynamical equations are chaotic~\cite{kostuk}. The equations for the states $x_a(t);\; a = 1, 2, ..., D$ are meant to describe `weather stations' on a periodic spatial lattice. This model is widely used in atmospheric science as a testbed for the exploration of innovative data assimilation ideas.

Our example selects D = 11, and displays the action level plot for L = 2, 4, 5, and 6 observations at each measurement time within the window $[t_0,t_F]$. We perform a `twin experiment' wherein we generate $D$ time series $\{x_a(t);\;a=1,2,...,D\}$ for Eq. (\ref{lorenz96}) using a standard adaptive fourth order Runge-Kutta algorithm with a time step $\Delta t = 0.025$ and an initial condition $\x(t_0)$ drawn from a uniform distribution over the range of the variables $\x(t)$, namely [-10, +10]. To these solutions of Eq. (\ref{lorenz96}) we add Gaussian noise with mean zero and variance $\sigma^2 = 0.2$ to each time series $x_a(t)$. These noisy versions of our model time series constitute our `data' $\{y_a(t)\}$.  $L$ of these D time series are presented to the model at times $\tau_n;\; t_0 \le \tau_n \le t_F$.

The measurement window is from $t_0 = 0$ to $ t_F = 4.125$. $L = 2, 4, 5, 6$ `measurements' are made at each time step; these are the $\y(\tau_n)$. The measurement error matrix $\R_m$ is taken to have diagonal elements at each measurement time $\tau_n$ and is zero at other times. Its magnitude is taken as $R_m = 1/\sigma^2 = 5$. 

The model error matrix $R_f(a)$ is also taken as diagonal, with elements along the diagonal $R_f = R_{f0} 2^{\beta}$, in performing the VA procedure, and we take $\beta = 0,1,2,\dots\,$. $R_{f0}$ was chosen 0.01. 

The minimizations of nonlinear objective functions in the example using the Lorenz96 model was performed using the public domain software IPOPT~\cite{ipopt} with a front end script written in Python.

\begin{figure}[htbp] % float placement: (h)ere, page (t)op, page (b)ottom, other (p)age
  \centering
  % file name: E:/machines_da/lor96d11data/lor96d11l4.jpg
%  \includegraphics[width=3.74in,height=2.67in,keepaspectratio]{lor96d11l4}
%2461,width=2.67in,height=3.74in,keepaspectratio]{actionlevelslor96d11l2new.jpg}
  \includegraphics[width=2.67in,height=3.74in,keepaspectratio]{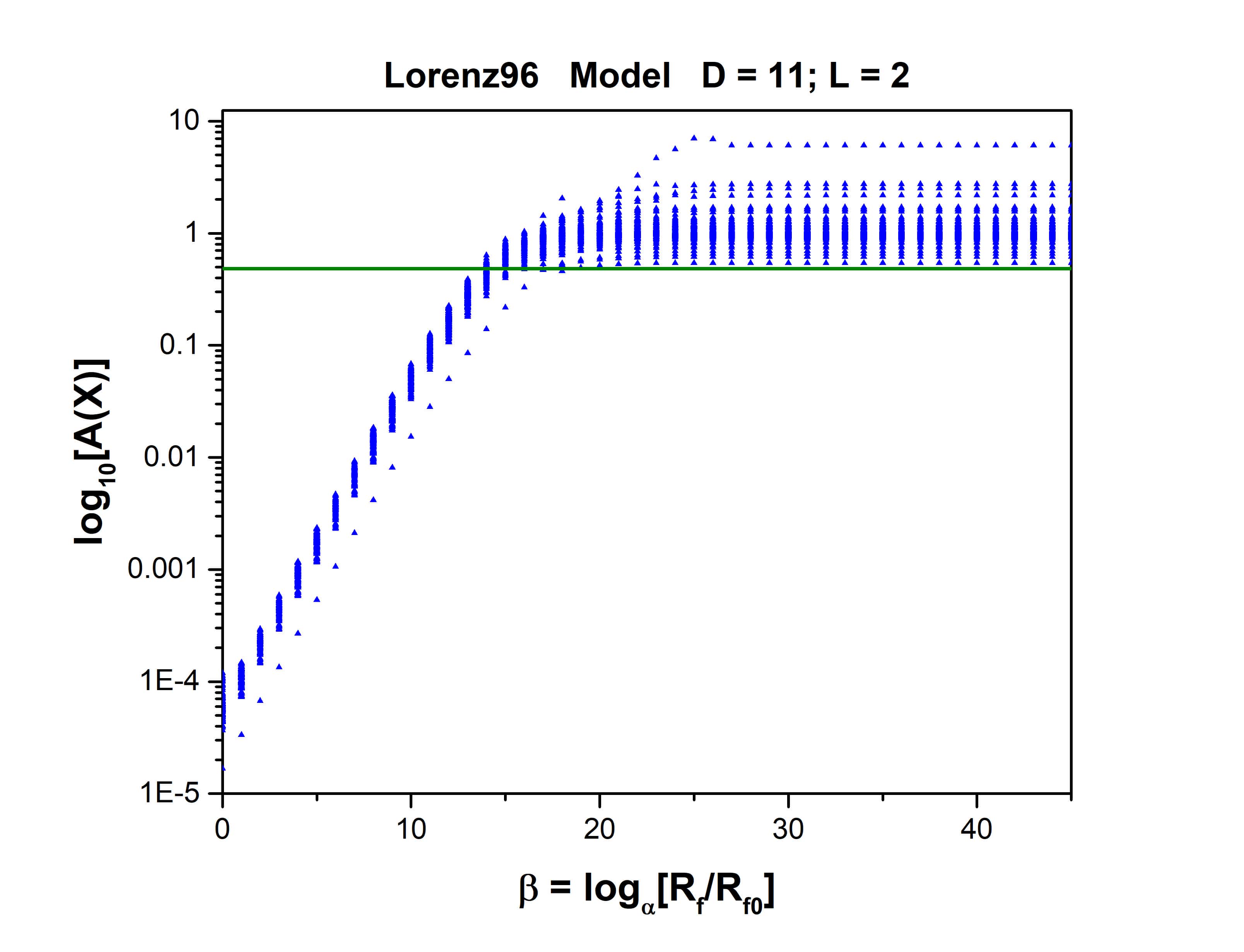}
  \includegraphics[width=2.67in,height=3.74in,keepaspectratio]{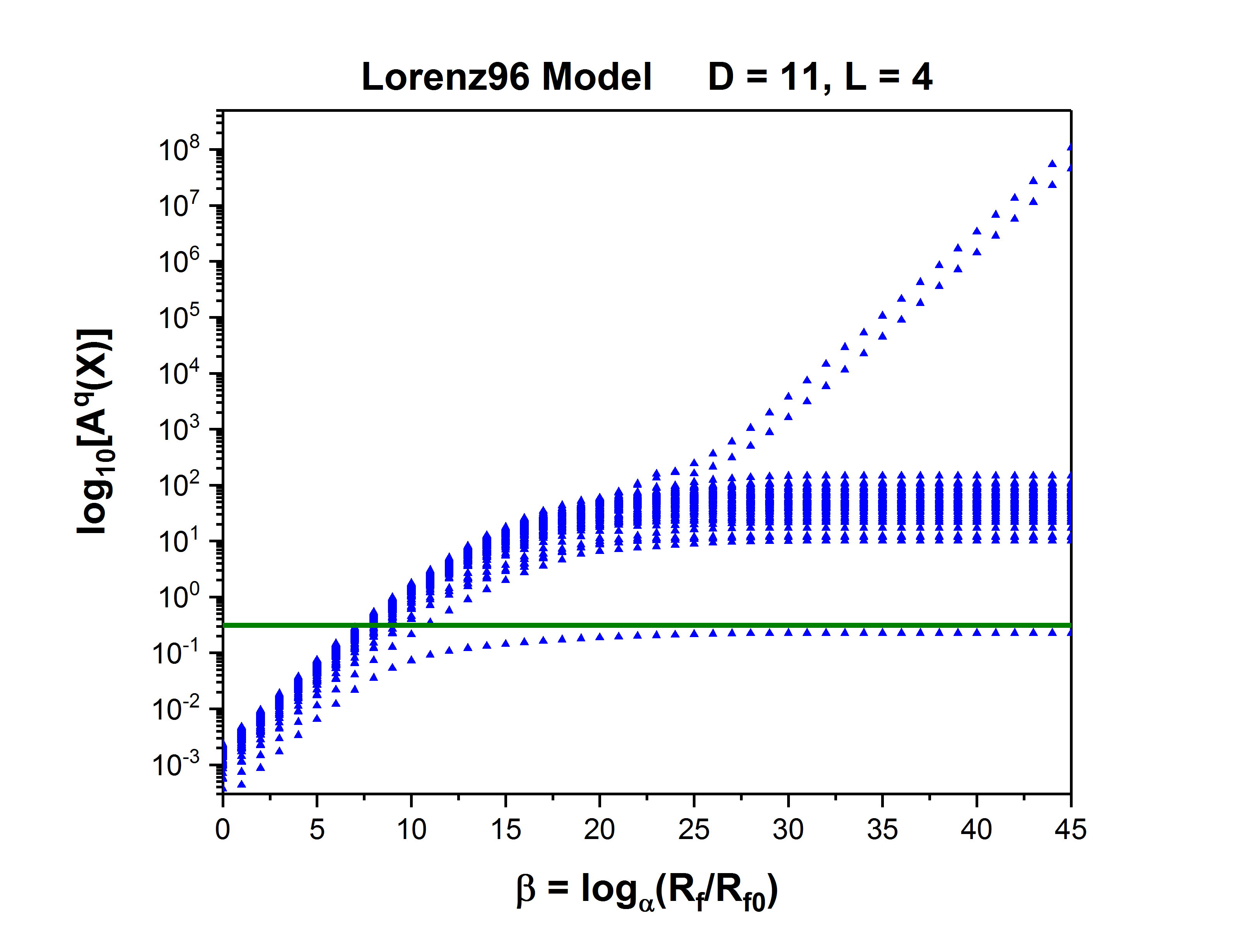}
  \includegraphics[width=2.67in,height=3.74in,keepaspectratio]{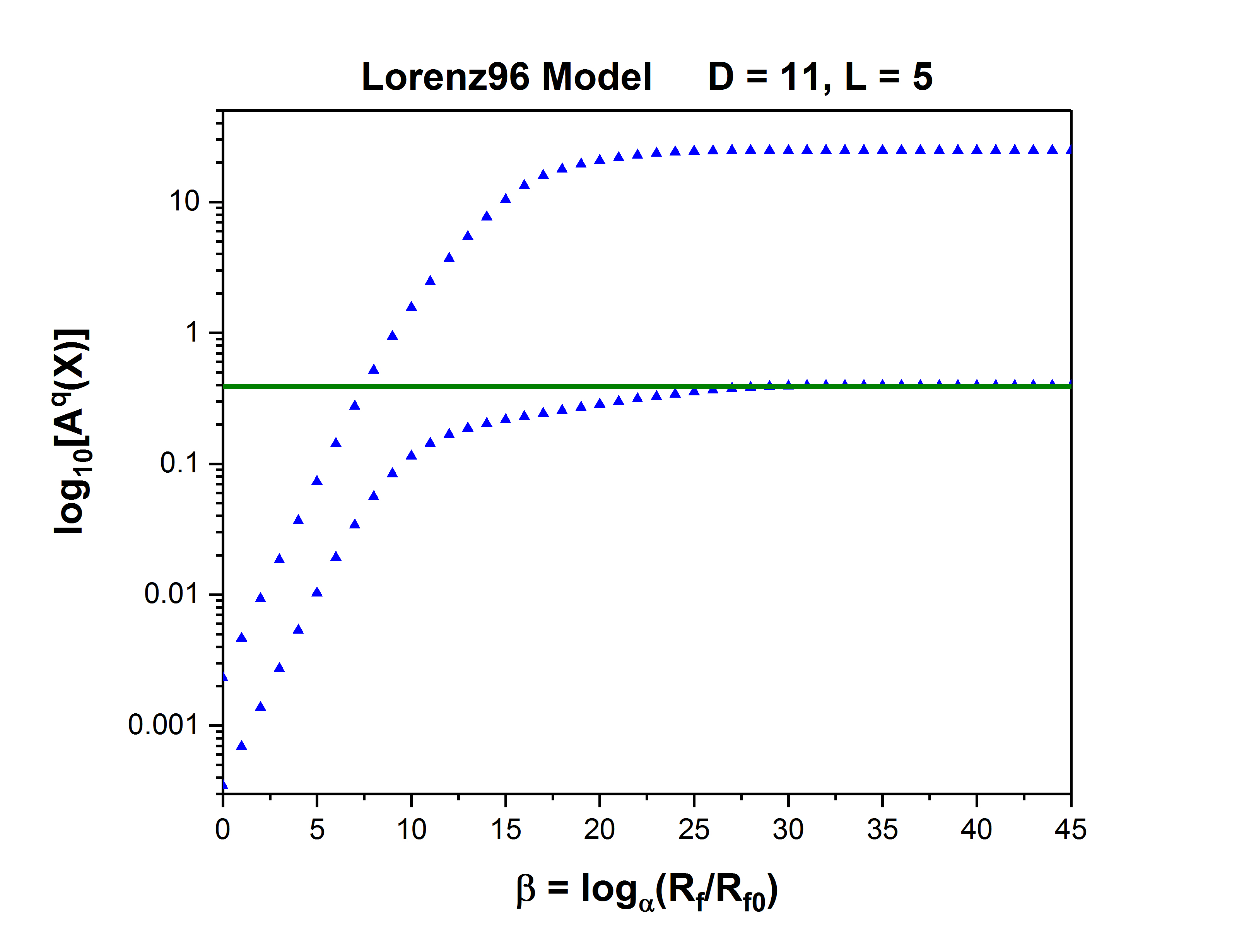}
  \includegraphics[width=2.67in,height=3.74in,keepaspectratio]{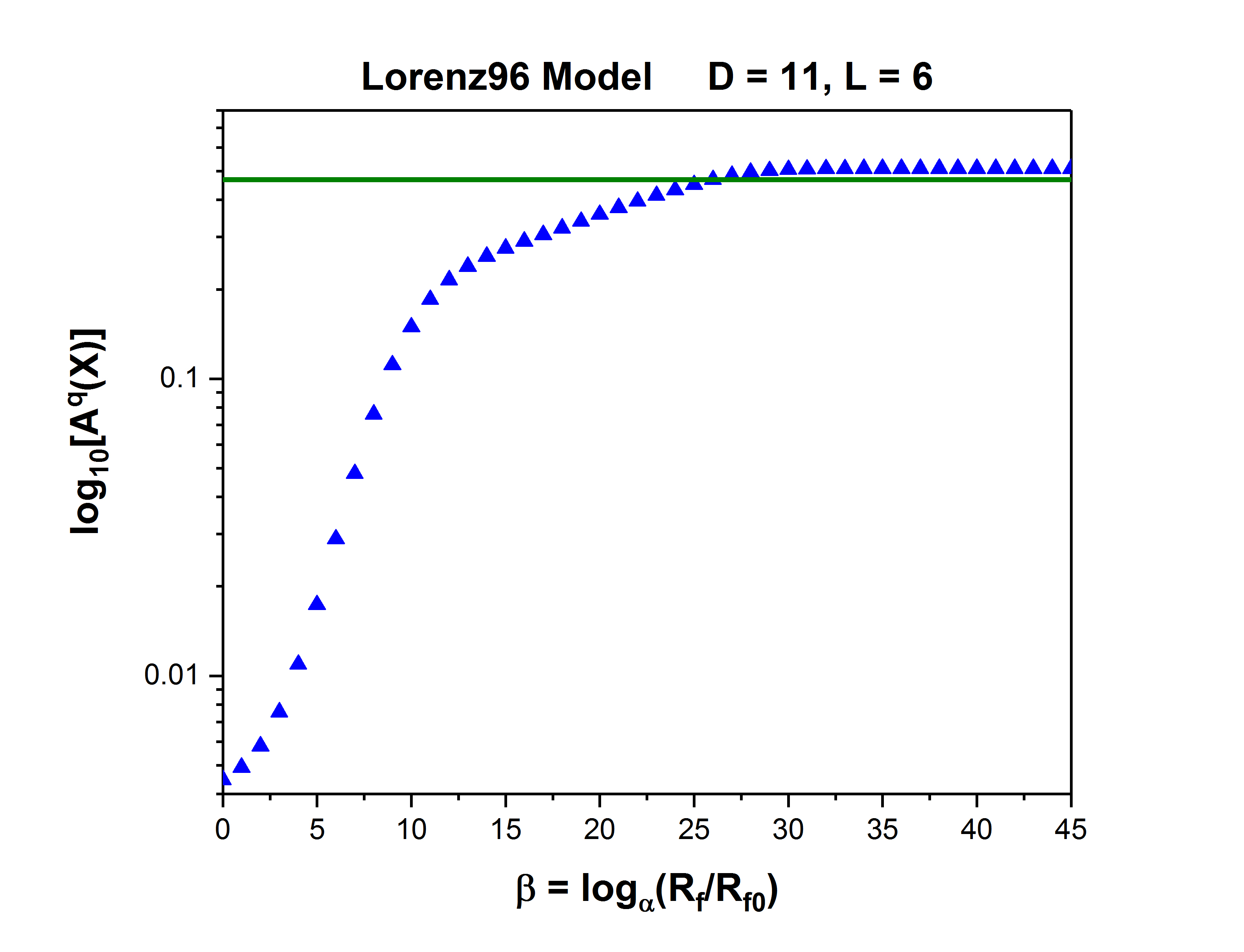}
  \caption{Action level plots for the Lorenz96 model Eq. (\ref{lorenz96}) with D = 11 and $\nu = 10.0$. The variational annealing procedure is performed with 100 different initial conditions for the minimization of the action at each value of $R_f = R_{f0}\alpha^{\beta};\;\beta = 0, 1, ...;\;\alpha = 2$. {\bf Top Left Panel}: L = 2 measurements at each measurement time. At L = 2, there are many minima, but none so much smaller than the others that it dominates the expected value integral Eq. (\ref{expected0}). {\bf Top Right Panel}:  L = 4 measurements at each measurement time. At L = 4, the action in path space $\X$ has numerous local minima. The lowest minimum has an action value much smaller than the action values from the other minima, and this dominates the expected value integral Eq. (\ref{expected0}). {\bf Bottom Left Panel}:  L = 5 measurements at each measurement time.  At L = 5, the number of minima found is only two, and again the lowest minimum dominates the expected value integral. {\bf Bottom Right Panel} L = 6 measurements at each measurement time.   At L = 6, there is only one minimum of the action. 
The solid green line is the expected value of the measurement error term. This is distributed as $\chi^2$. As the action becomes independent of $R_f$, its expected value should equal this value.}
  \label{lor96d11l4}
\end{figure}

In Fig. (\ref{lor96d11l4}) we display action level plots for $L = 2, 3, 4,$ and $6$ observations at each measurement time. As we can see in the ${\bf Top Left Panel}$, where $L = 2$,  there are numerous local minima in the action $A_0(\X)$ for all values of $R_f \ge R_{f0}$, and these remain to large $R_f$. None of these minima is very far separated from the paths with the smallest minimum, so that the evaluation of the expected value integrals Eq. (\ref{expected0}) would require contributions from many maxima of the conditional probability distribution. 

When $L = 4, {\bf Top Right Panel}$ we begin to see an isolated action level whose contribution to the expected value integral is overwhelmingly larger than the contribution from path giving rise to the next largest action level. The value $L = 4$ is consistent with the observation in~\cite{kostuk} that around $0.4 D$ the instabilities in the Lorenz96 state space appear to be controlled by the data assimilation process. 

At $L = 5$ or $6$, ${\bf Bottom Panels}$, we see that the dominance of the lowest action level is even clearer. The horizontal olive colored line is the expected value of the measurement error term in the action. This is a sign of the consistency of the data assimilation calculations.

\begin{figure}[tbph] % float placement: (h)ere, page (t)op, page (b)ottom, other (p)age
  \centering
  % file name: E:/machines_da/lor96d11data/paramlord11l6blowup.jpg
%  \includegraphics[width=2.67in,height=3.74in,keepaspectratio]{actionlor96d11l6}
%  \includegraphics[width=2.67in,height=3.74in,keepaspectratio]{actionlor96d11l7}
%  \includegraphics[width=2.67in,height=3.74in,keepaspectratio]{lor96d11L6modelsteps}
%2461,width=2.67in,height=3.74in,keepaspectratio]{lor96d11l6levels.jpg}
%2461,width=2.67in,height=3.74in,keepaspectratio]{paramlord11l6100steps.jpg}
  \includegraphics[width=2.67in,height=3.74in,keepaspectratio]{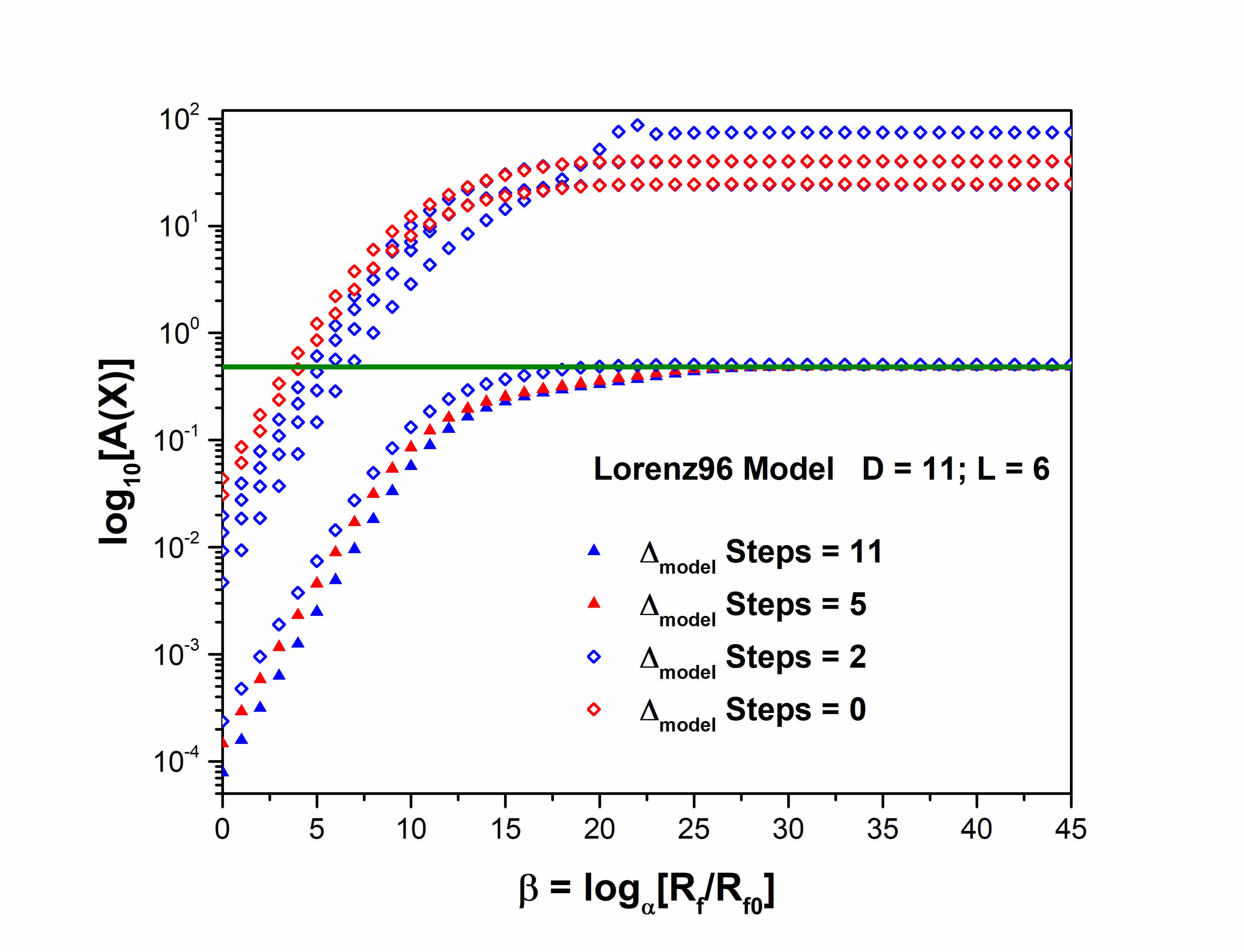}
  \includegraphics[width=2.67in,height=3.74in,keepaspectratio]{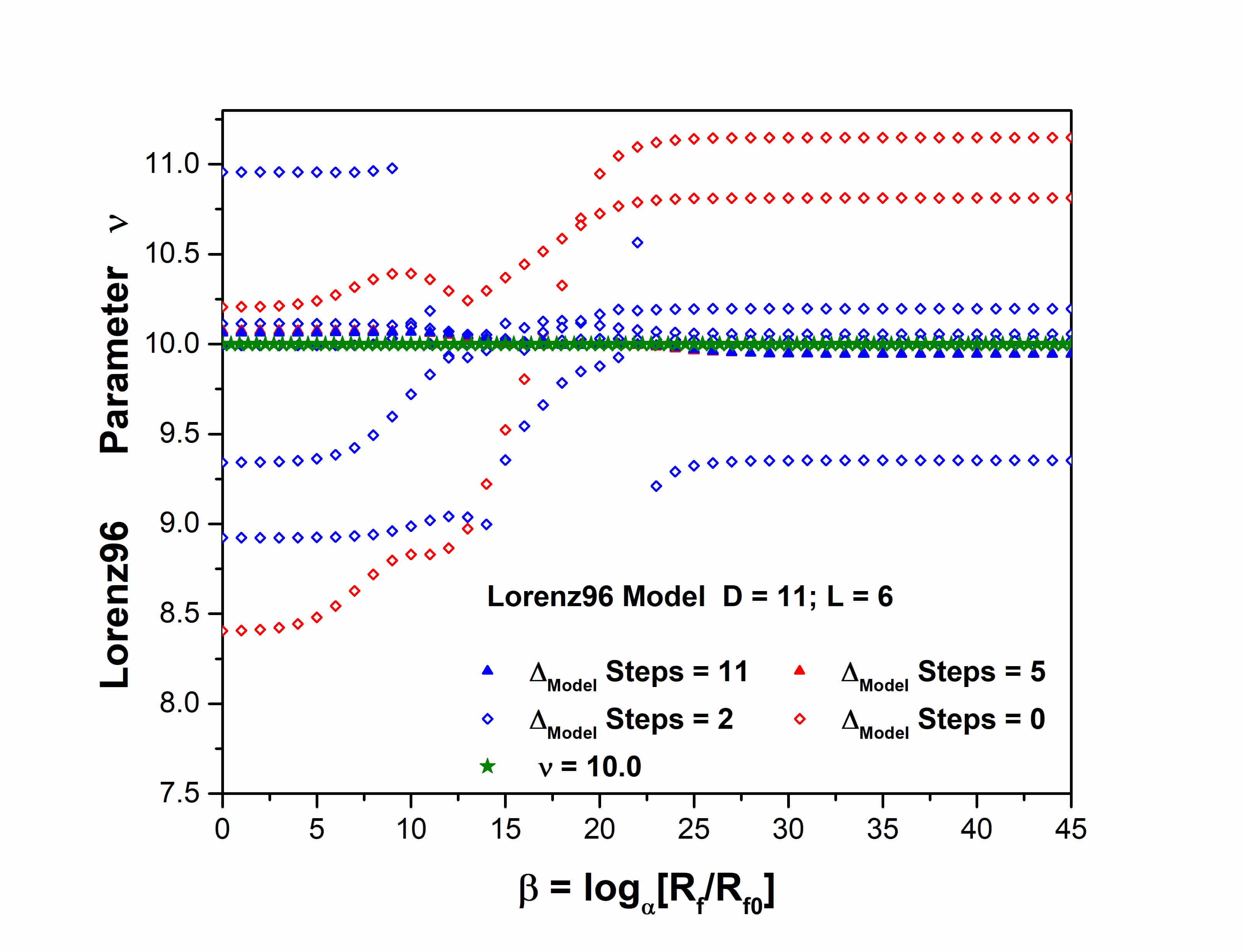}
  \caption{Parameter estimation  and action level results for the Lorenz96 model, D = 11, L = 6. The parameter value $\nu = 10.0$ was used in the twin experiments on the model. Observations were made every $\Delta t_{obs} = 0.15$, and L = 6 measurements were made at each observation time. {\bf Left Panel}: {\tt Action Level Estimates} These are the action levels when $L = 6$ observations are made at each measurement time and with the choice of 0, 2, 5, and 11 model evaluation steps between measurement times. The horizontal olive green line indicates the expected action level for large $R_f$. {\bf Right Panel} {\tt Parameter estimates} Between the observations the model was evaluated 0, 2, 5 and 11 times leading to $\Delta t_{model}$ = 1.0, 0.33, 0.16, and 0.08 $*\Delta t_{obs}$. The parameter estimates are quite accurate for 5 and for 11  model time steps between observations. They are more distributed for 0 or 2 model step between observations. One can associate the idea of increasing the number of model steps between observations as equivalent to `deepening' the hidden (unobserved) layers. The horizontal olive green line indicates the parameter value, 10.0, used in generating the data.}
%{\bf Top Right Panel} {\tt Action Level Estimates; L = 7} The action levels when $L = 7$ observations at each measurement time and with the choice of 0, 1, 2, 5, and 11 model evaluation steps between measurement times. }
  \label{actionparamd11l6steps}
\end{figure}

In Fig. (\ref{actionparamd11l6steps}) we explore another aspect of the action level plots. We still use $D = 11$, and we hold $L = 6$ fixed. The number of observations within the window $[t_0, t_F]$ has been reduced from 165 to 28 and we move the model forward between observations 0, 2, 5 or 11 times. This is to provide an analogy to how many layers are present in an equivalent machine learning example. Our example here differs by having many entry points in the measurement window while the machine learning example has only one. We display in the ${\bf Left Panel}$ the action level plots for the selected number of model evaluation steps. As one can see for 0 and 2 intermediate steps we have many persisting minima of the action. At 5 and 11 intermediate steps, there is only a single minimum that is found, and for large $R_f$ it comes to the same action level as with 2 intermediate steps. All are consistent with the expected value of the measurement error term. This calculation, performed in a machine learning context provides information on how many hidden layers are required to achieve a desired accuracy.

In the ${\bf Right Panel}$ we display the accuracy of the estimation of the single parameter in the Lorenz96 model. It has been set at $\nu = 10.0$ in producing the data, and that value is clearly selected for 5 or 11 intermediate model evaluations, while it is not so clearly selected for 2 intermediate steps and with zero intermediate steps there is a consistent few percent error.

%fig 1

%fig 2

We see, in this collection of calculations, as noted earlier~\cite{ye2014precision,ye2015physrev}, the ability to identify the dominant minimum of the action depends on the number of measurements presented during the statistical data assimilation procedure embodying transfer of information from the data to the model. In data assimilation this is associated with the number of positive conditional Lyapunov exponents~\cite{kostuk} of the model. In the machine learning instantiation it may play the same role when the number of data presented at the output layer is not sufficient to determine the parameters and hidden states in each layer.

We also see the analogue of deepening the network produces higher accuracy estimates of conditional expected values.

\subsection{A Multi-Layer Perceptron; Feedforward Network}

We constructed a feedforward network with $l_F$ layers: One input layer at $l_0$ and one output layer at $l_F$. This network has $l_F - 2$ hidden layers. We analyzed $l_F = 20, 30, 50$ and $100$. There are $N = 10$ `neurons' in each layer. The activity in neuron $j$ in layer $l$ $x_j(l)$ is related to $x_j(l-1)$ in the previous layer as
\be 
x_j(l) = g(W(l)\x(l-1));\;\;g(z) = 0.5[1 + \tanh(\frac{z}{2})].
\ee 
%The weights are drawn from $u[-1,1]$ and input layer values are between 0 and 1. 
%The number of known values of $\x(l_0)$ is called L. 
We also investigated the ``ReLU"-like function $g(z) = \log[1+ e^z]$, but we do not report on those results.
The activations at the input layer are drawn from a Gaussian $N(0,1)$. The weights are selected from a uniform distribution $U[-0.1,0.1]$. Gaussian measurement noise was added to the `data' generated by the model; this has zero mean and variance $0.0025$. 

Our data are for the twin experiment, where we know all weights, all inputs and generate all values of $\x(l)$ in each layer $[l_0,l_F]$. These are recorded, and Gaussian noise with mean zero and variance 0.0025 is added to the data at layers $l_0 = 1$ and $l_F$ starting from known values $\x(l_0)$. These data are $y_i(1)$ for the input layer and $y_i(20)$ for the output layer.

$M$ input/output pairs are presented to the model with L = 1, 5, and 10 inputs $y^{(k)}_i(l_0)$ at layer $l_0$ and L = 1, 5, 10 data outputs $y^{(k)}_i(l_F)$ at layer $l_F$. k = 1,2,...,M, and we investigated $M = 1, 10, 100$.

We minimize the action over all the weights and the states $x^{(k)}_a(l)$ at all layers of the model:
\bea 
&&A_M(\X) = \frac{1}{M} \sum_{k =1}^{M}\biggl\{ \frac{R_m}{2L}\sum_{r=1}^L\, \biggl[ (x^{(k)}_r(l_0) - y^{(k)}_r(l_{0}))^2 + (x^{(k)}_i(l_F) - y^{(k)}_i(l_F))^2\biggr]  \nonumber \\
&&+ \frac{R_f}{N(l_F-1)} \sum_{l=1}^{l_F -1} \sum_{a=1}^N \biggl[ x^{(k)}_a(l+1) - g(W(l)\x^{(k)}(l)) \biggr]^2\biggr \},
\eea
where we have N = 10 neurons in each layer and $L \le N$ data at the input $l_0$ and at the output layers $l_F$. 

We use the variational annealing procedure described above to identify the action levels for various paths through the network. The initial value of $R_{f0}/R_m$ is taken to be 10$^{-8}$ and this is incremented via $R_f/R_m = R_{f0} \alpha^{\beta}$ with $\alpha = 1.1$ and $\beta = 0, 1, ..., $. 
%In Figure (\ref{mlpactionM10}) we show the action level plots for L = 1 and 5 at $M = 10$, and in Figure (\ref{mlpactionM100}) we show the action levels for $L = 1, 5, 10$ for $M = 100$.

In the numerical optimizations for the machine learning example we used  L-BFGS-B~\cite{byrd,zhu}.

There are at least two ways to present more information to the model in this setting:
\begin{itemize}
  \item increase the number of training pairs available to  the network at $l_0$ and $l_F$; this is our number $M$. $M$ can be chosen as large as the user wishes.
  \item increase the number of components of the input/output pair vectors; this is our number $L$. $L \le N$, the number of neurons in $l_0$ or $l_F$.
\end{itemize} 
The resolution of the model in its ability to capture variations in activity from layer to layer is improved by increasing the number of layers $l_F$.

\begin{figure}[htbp] % float placement: (h)ere, page (t)op, page (b)ottom, other (p)age
  \centering
  % file name: E:/machines_da/machine_pr_data/data_062717/actionlevelslF20M100L10.tif
% 2461,width=2.67in,height=3.74in,keepaspectratio]{actionlevelslF20M100L10.jpg}
% 2461,width=2.67in,height=3.74in,keepaspectratio]{actionlevelslF30M100L10.jpg}
% 2461,width=2.67in,height=3.74in,keepaspectratio]{actionlevelslF50M100L10.jpg}
%2461,width=2.67in,height=3.74in,keepaspectratio]{actionlevelslF100M100L10.jpg}
  \includegraphics[width=2.67in,height=3.74in,keepaspectratio]{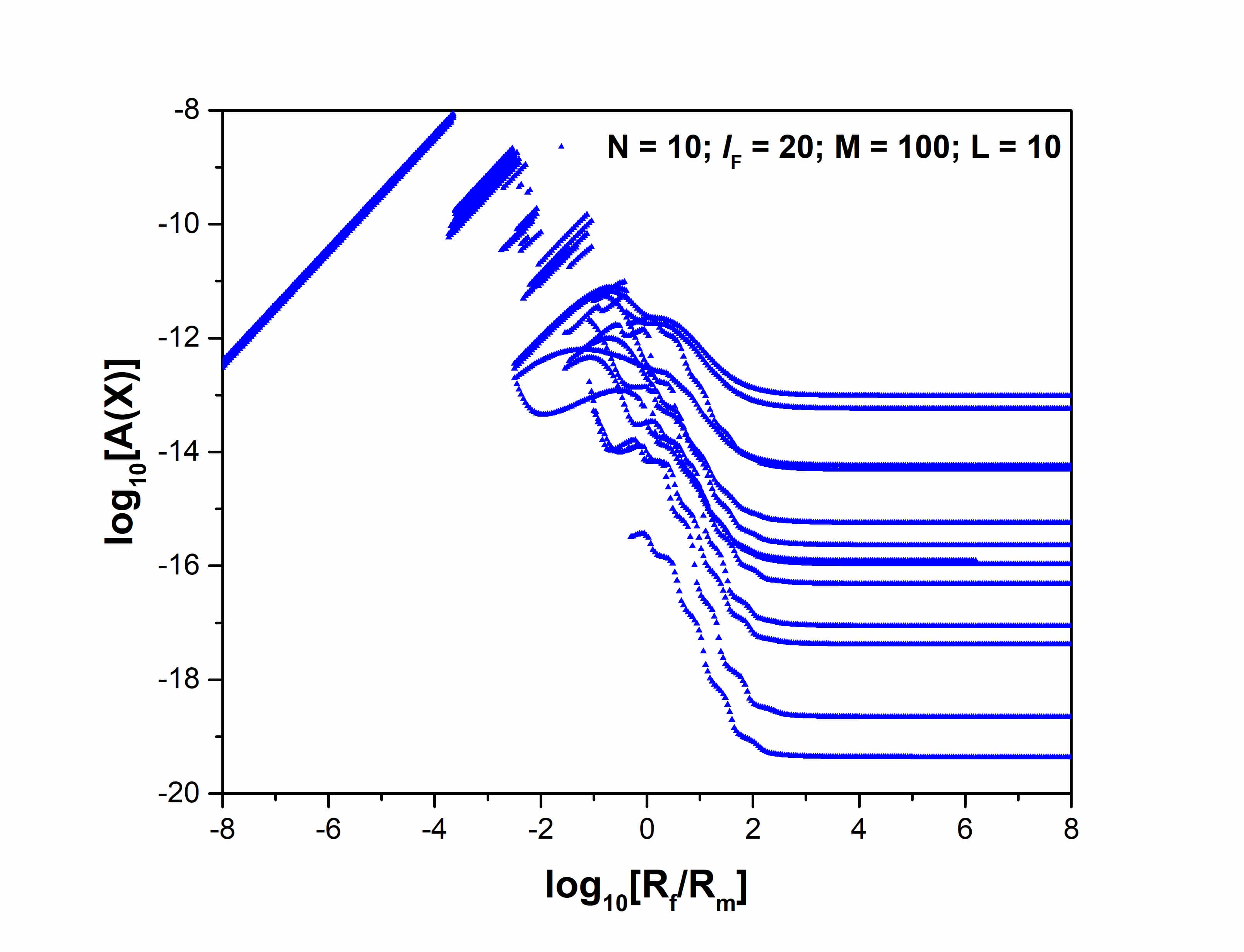}
  \includegraphics[width=2.67in,height=3.74in,keepaspectratio]{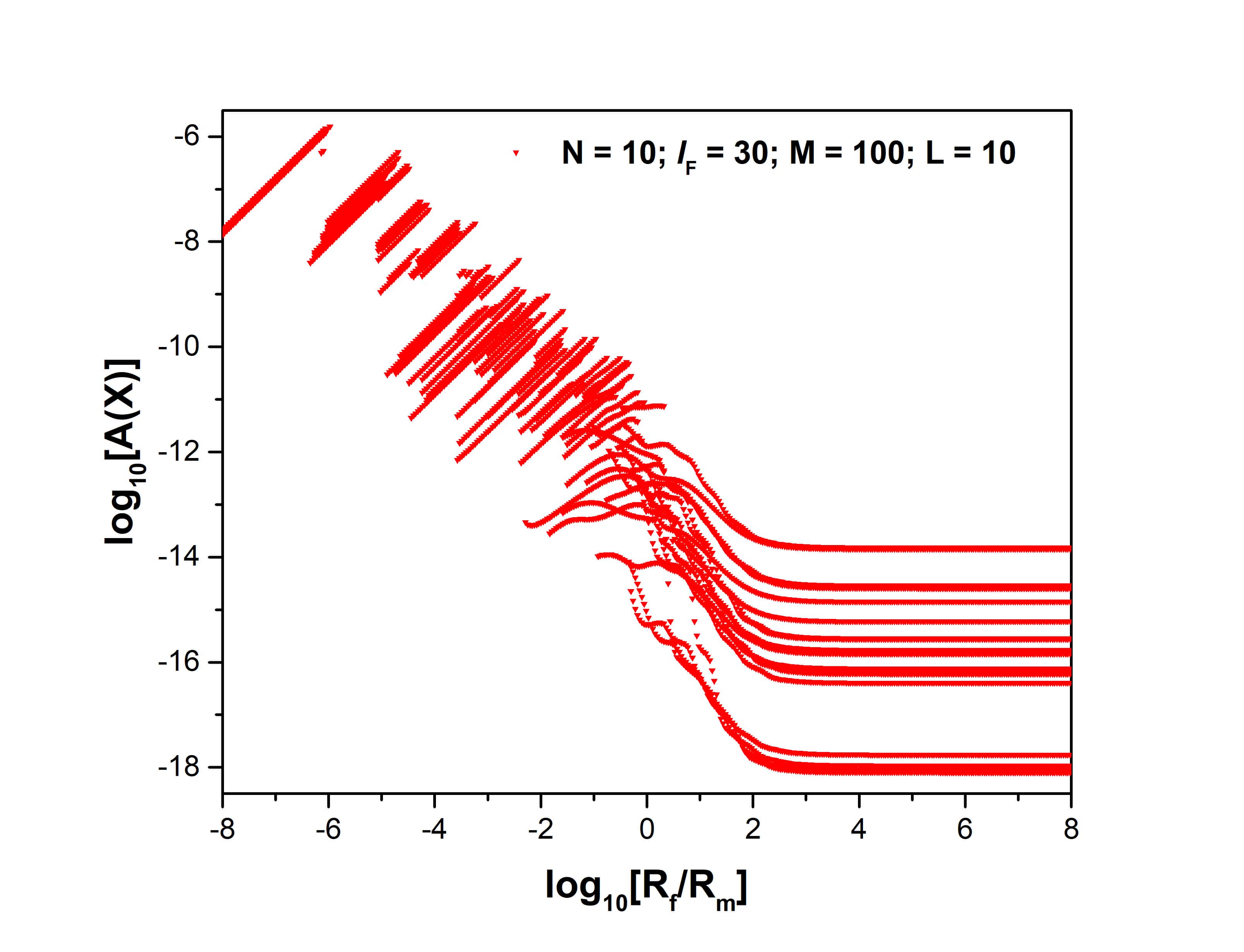}
  \includegraphics[width=2.67in,height=3.74in,keepaspectratio]{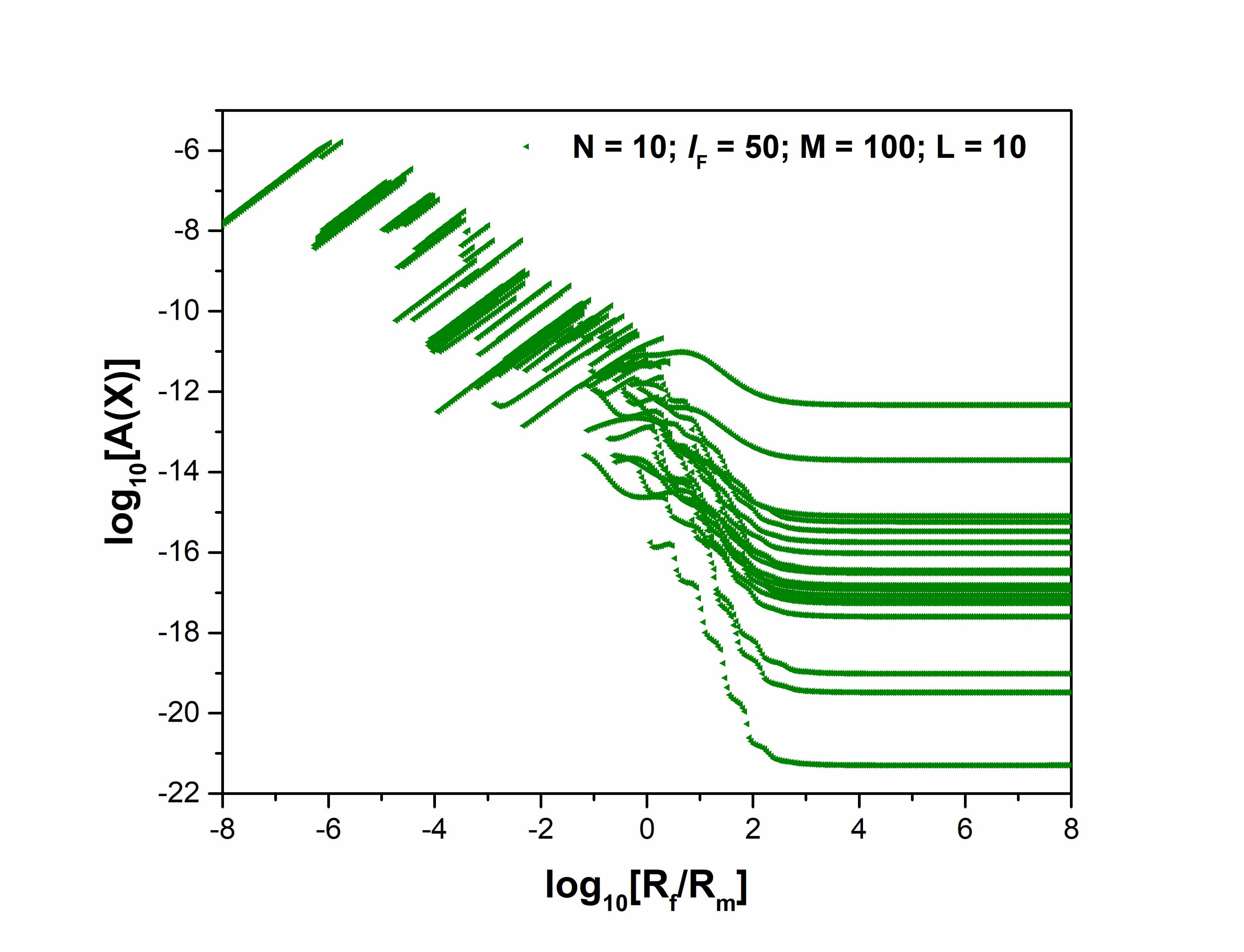}
  \includegraphics[width=2.67in,height=3.74in,keepaspectratio]{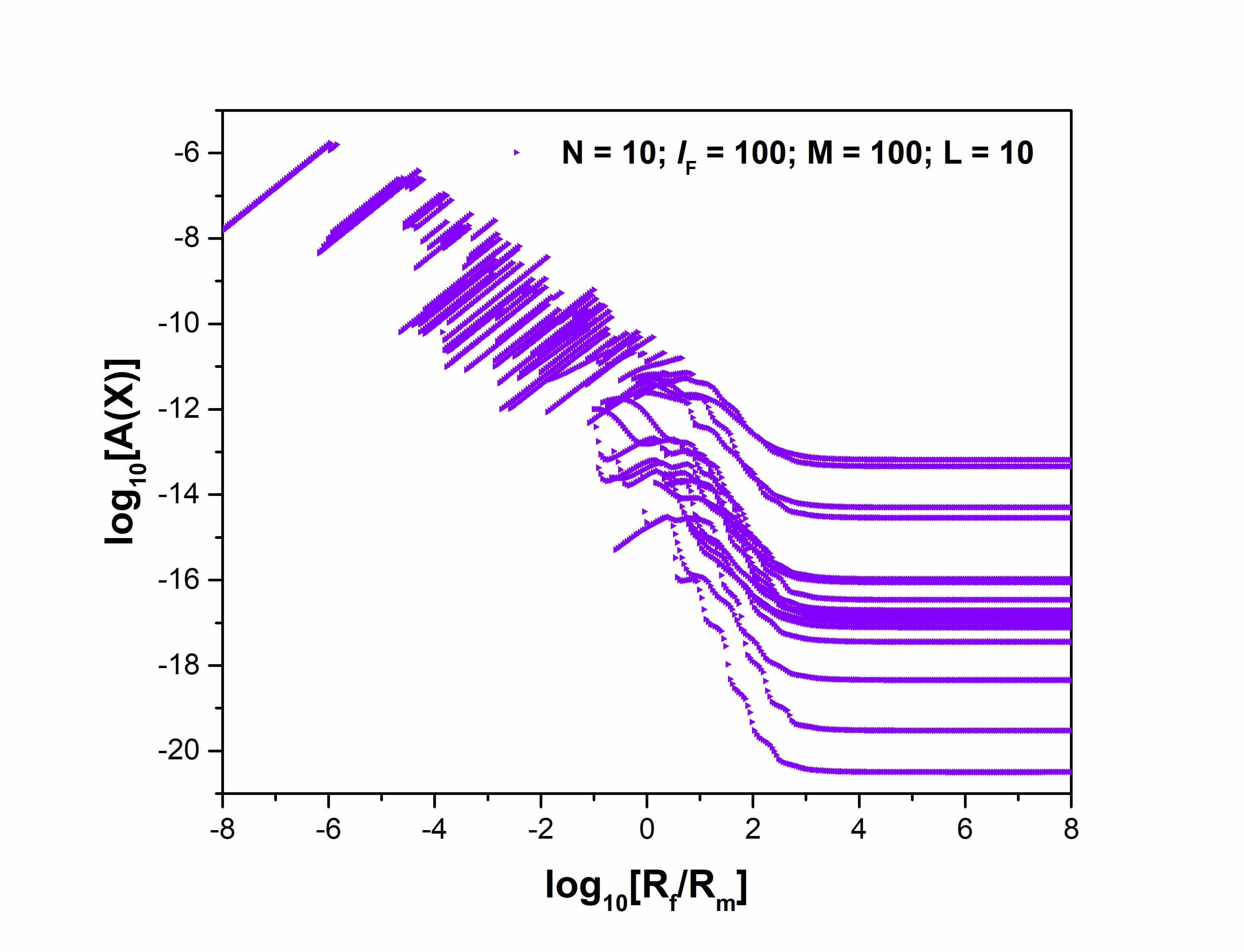}
  \caption{Holding the number of neurons in each layer fixed at 10, the number of input/output pairs fixed at $M = 100$, and the number of inputs and outputs at $l_0$ and $l_F$ fixed at $L = 10$, we vary the number of layers (the deepening of the network) and examine the action level plots arising from the variational annealing procedure.  {\bf Upper Left Panel} $l_F = 20$ {\bf Top Right Panel} $l_F = 30$ {\bf Bottom Left Panel} $l_F = 50$ and {\bf Bottom Right Panel} $l_F = 100$.}
  \label{actionlevelslF20M100L10}
\end{figure}

In Fig. (\ref{actionlevelslF20M100L10}) we explore the action levels as a function of $R_f/R_m$ as we increase the number of layers $l_F$ in the model: $l_F = 20,30,50,100$. We hold fixed the number of neurons $N = 10$, the number of training pairs $M = 100$ and the number of inputs and outputs $L = 10$ at $l_0$ and $l_F$. 

In each case there are many local minima, but only when $l_f = 50$ does the lowest action minimum significantly split from the other action minima and qualify to dominate the expected value integral Eq. (\ref{expected0}). When $l_F$ is increased from 50 to 100, the lowest action minimum comes closer to the second lowest minimum. This is seen as a result of the much larger number of weights to be estimated at the latter value of $l_F$ while we are holding fixed through the values of $L$ and $M$ the information available to make those estimations.
%first option: increasing $L$ at a fixed $M$. We see i the display that the action level plot for $M = 10$ and both $L = 1$ and $L = 5$ are quite cluttered with local minima, and no minimum stands out as dominating the expected value integral. We do not display the outcome of increasing $L$ to 10, the maximum available in this network, as nothing really changes.

\begin{figure}[htbp] % float placement: (h)ere, page (t)op, page (b)ottom, other (p)age
  \centering
  % file name: E:/machines_da/machine_pr_data/data_062717/actionlevelslF50M100L1.tif
%2461,width=2.67in,height=3.74in,keepaspectratio]{actionlevelslF50M100L1.jpg}
%2461,width=2.67in,height=3.74in,keepaspectratio]{actionlevelslF50M100L5.jpg}
%2461,width=2.67in,height=3.74in,keepaspectratio]{actionlevelslF50M100L10.jpg}
  \includegraphics[width=2.67in,height=3.74in,keepaspectratio]{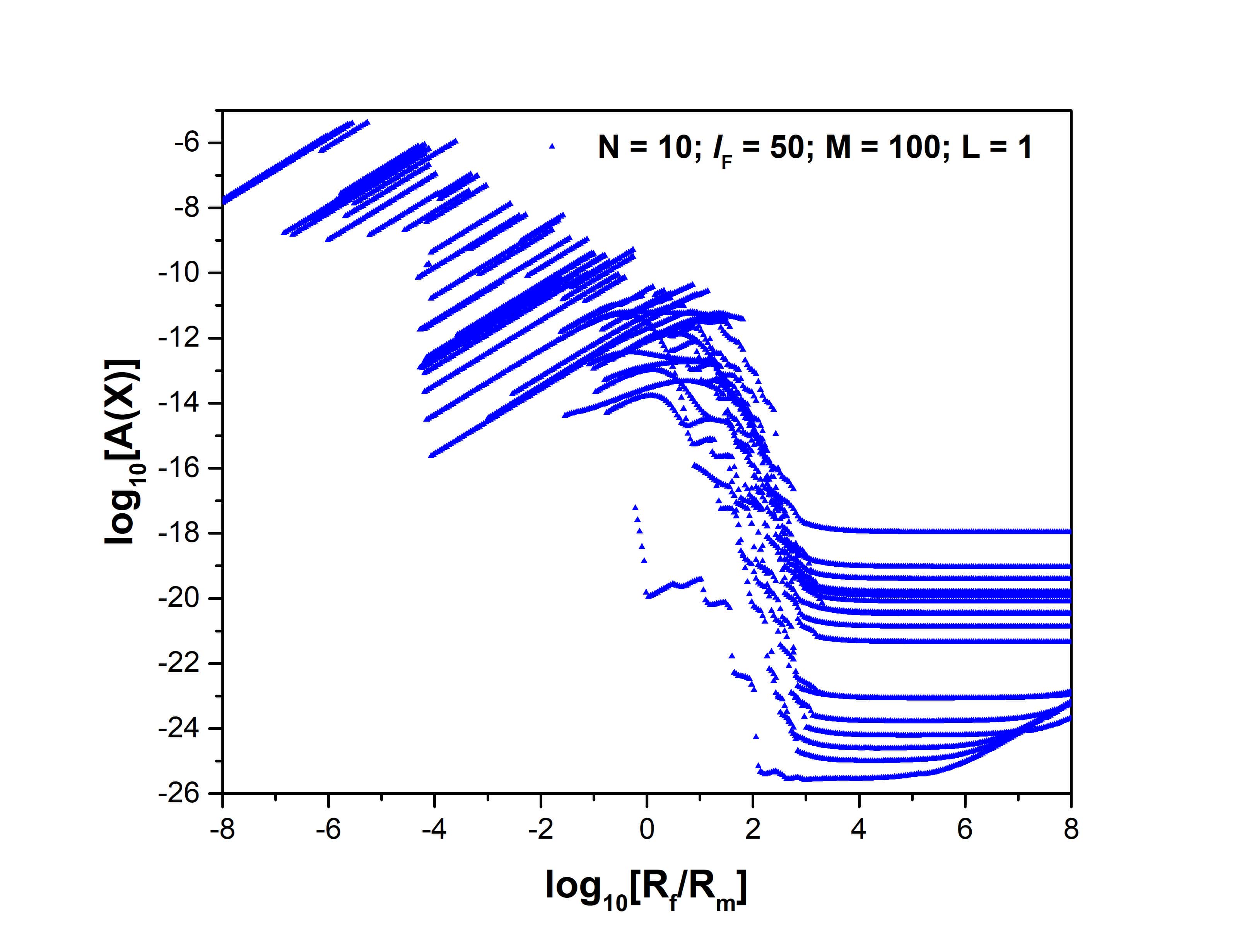}
  \includegraphics[width=2.67in,height=3.74in,keepaspectratio]{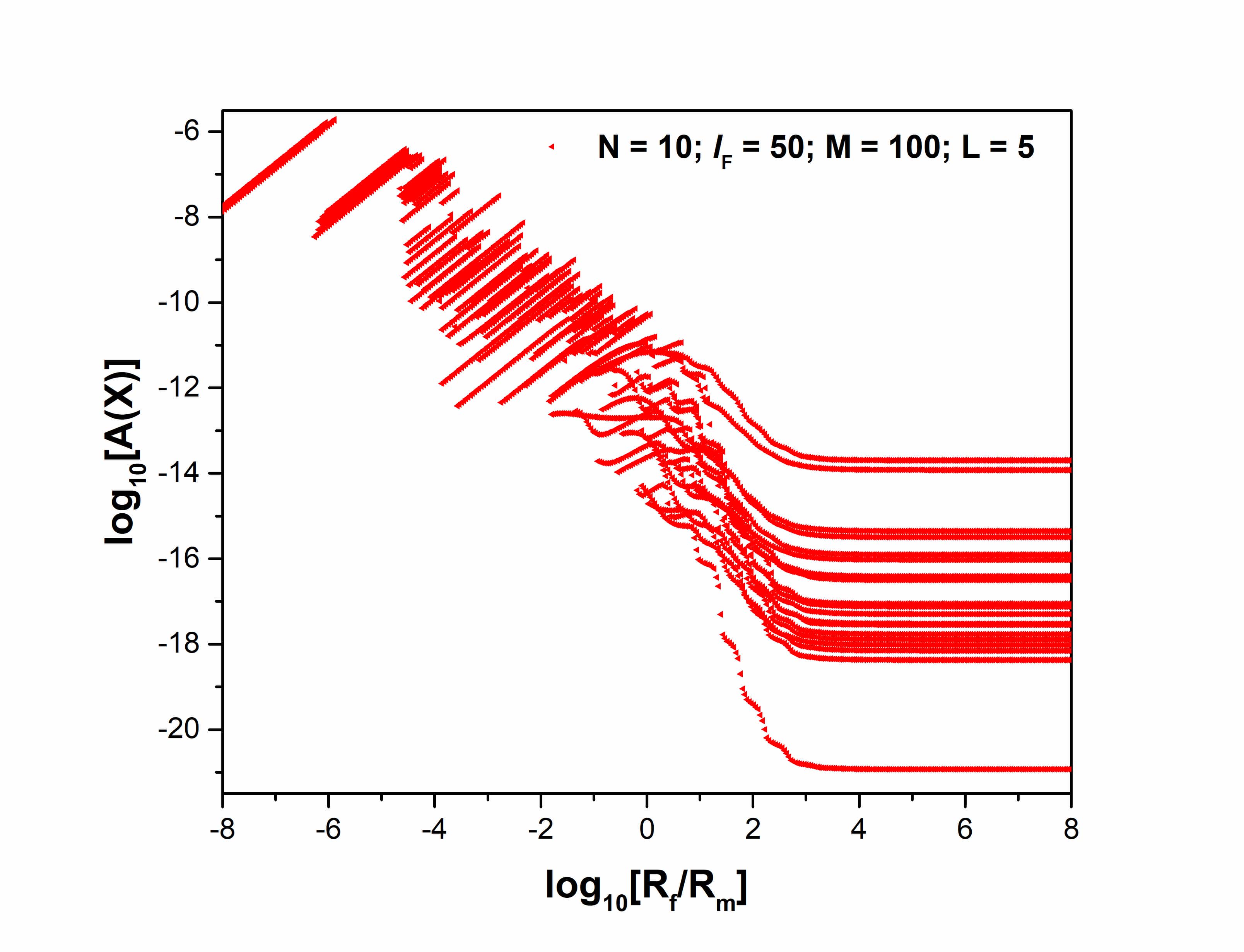}
  \includegraphics[width=2.67in,height=3.74in,keepaspectratio]{actionlevelslF50M100L10.jpg}
  \caption{Holding the number of neurons fixed $N = 10$, the number of layers fixed $l_F = 50$, and the number of input/output training pairs fixed $M = 100$, we display the action levels as we vary the number of inputs at $l_0$ and the number of outputs at $l_F$. {\bf Top Left Panel} $L = 1$. {\bf Top Right Panel} $L = 5$. {\bf Bottom Panel} $L = 10$.}
  \label{actionlevelslF50M100L1}
\end{figure}

In Fig. (\ref{actionlevelslF50M100L1}) we hold $M$ fixed at 100, and $l_F$ fixed at 50 while we look at $L = 1, 5, 10$ values of the dimension of input/output pairs. 

\begin{figure}[htbp] % float placement: (h)ere, page (t)op, page (b)ottom, other (p)age
  \centering
  % file name: E:/machines_da/machine_pr_data/data_062717/actionlevelslF50M1L10.tif
%2461,width=2.67in,height=3.74in,keepaspectratio]{actionlevelslF50M1L10.jpg}
%2461,width=2.67in,height=3.74in,keepaspectratio]{actionlevelslF50M10L10.jpg}
%2461,width=2.67in,height=3.74in,keepaspectratio]{actionlevelslF50M100L10.jpg}
  \includegraphics[width=2.67in,height=3.74in,keepaspectratio]{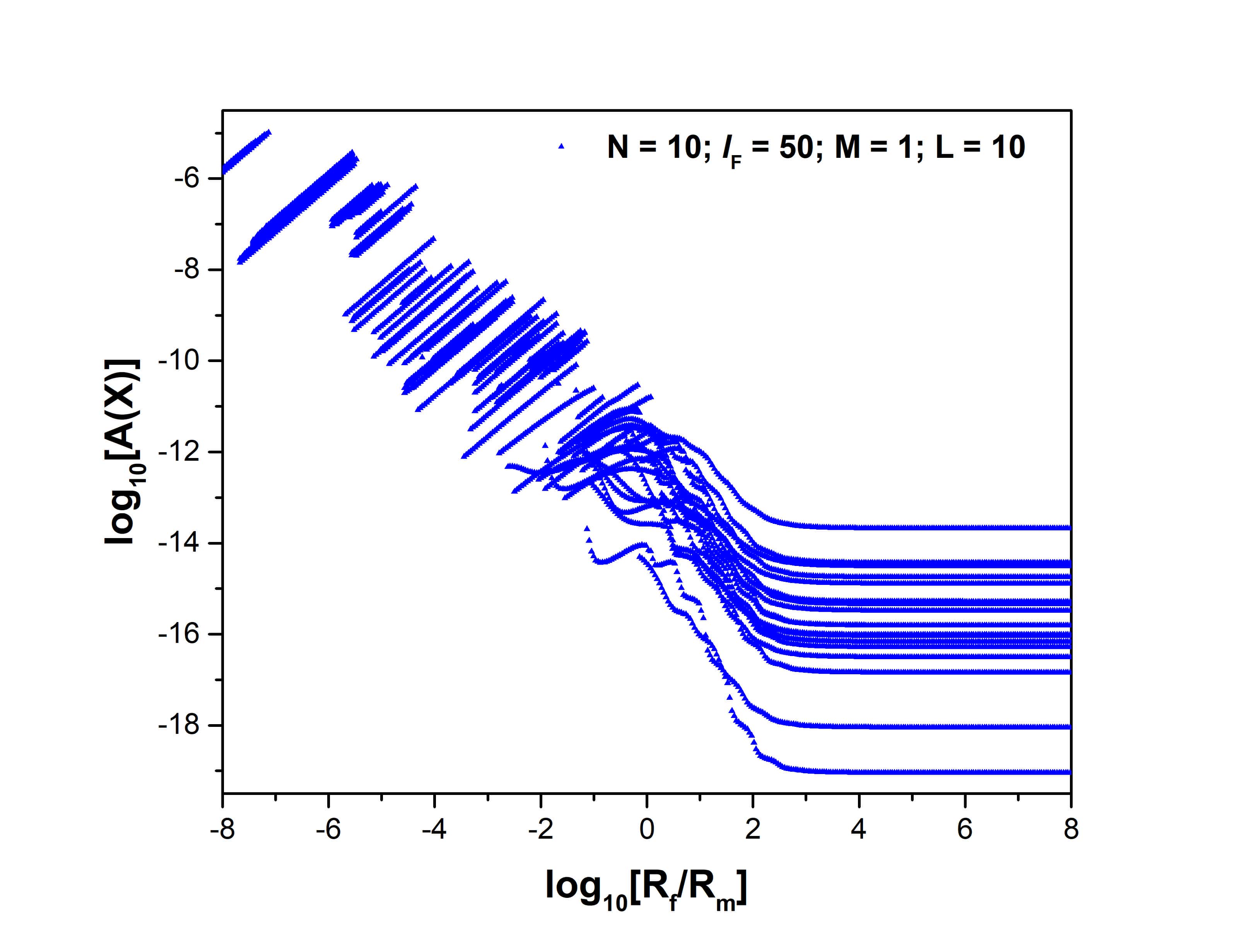}
  \includegraphics[width=2.67in,height=3.74in,keepaspectratio]{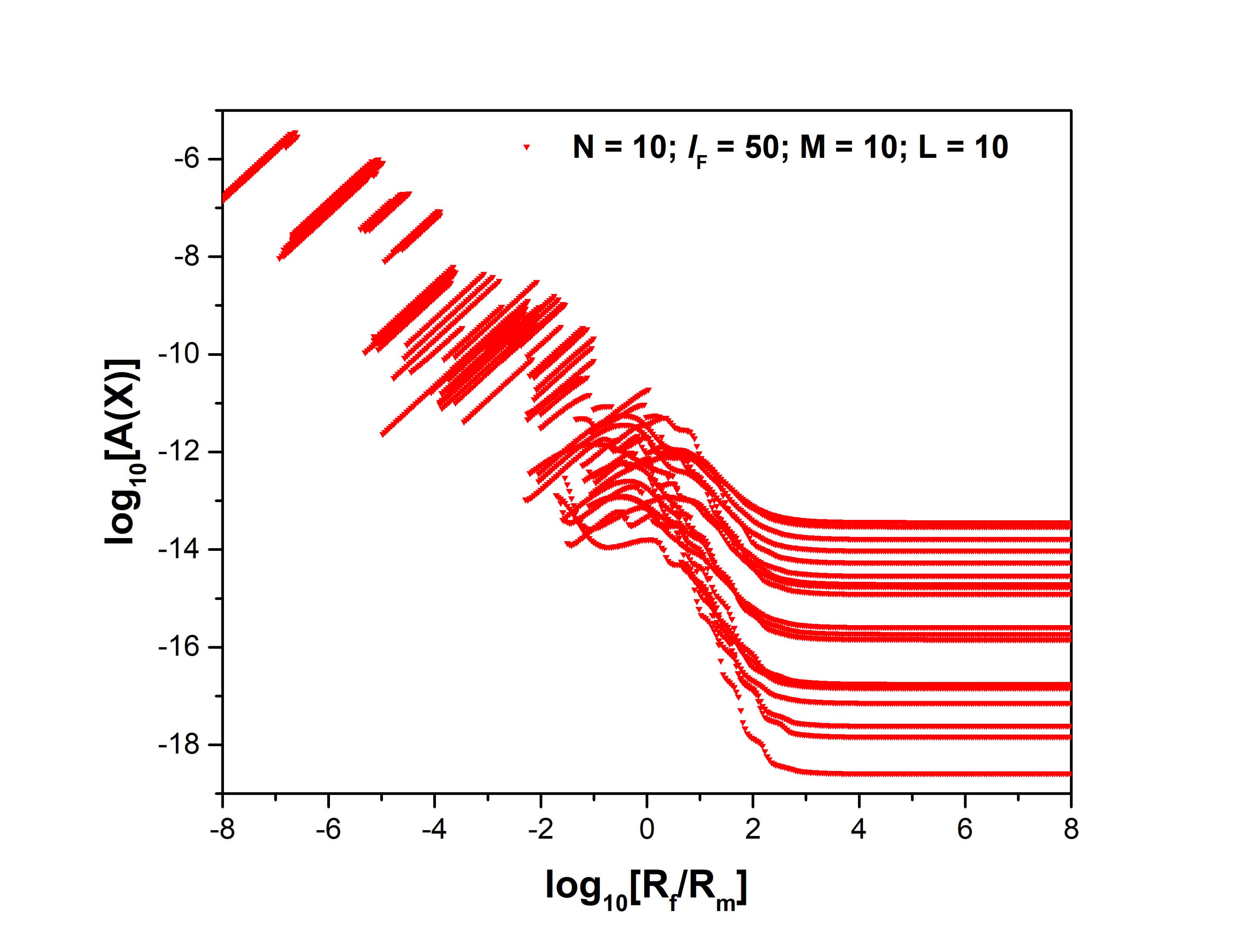}
  \includegraphics[width=2.67in,height=3.74in,keepaspectratio]{actionlevelslF50M100L10.jpg}
  \caption{Holding the number of neurons fixed $N = 10$, the number of layers fixed $l_F = 50$, and the number of inputs $L = 10$ at $l_0$ and the number of outputs at $l_F$, we display the action levels as we vary the number of training pairs $M$.  {\bf Top Left Panel} $M = 1$. {\bf Top Right Panel} $M = 10$. {\bf Bottom Panel} $M = 100$.}
  \label{actionlevelslF50M1L10}
\end{figure}

In Fig. (\ref{actionlevelslF50M1L10}) we hold fixed the number of layers $l_F$, the number of neurons in each layer $N = 10$ and the dimension of the input/output vectors $y^{(k)}_r(l_0)$ and $y^{(k)}_r(l_F); r = 1,2,..L = 10$. We show the effect of increasing the number of input/output pairs from $M = 1$ to $M = 10$ to $M = 100$. The emergence of a lowest action minimum as $M$ increases is displayed. This can serve as a candidate for approximating Eq. (\ref{expected0}). 

\begin{figure}[htbp] % float placement: (h)ere, page (t)op, page (b)ottom, other (p)age
  \centering
  % file name: E:/machines_da/machine_pr_data/data_062917/prederrorlevel0lF50L10M1_10.tif
%2461,width=2.67in,height=3.74in,keepaspectratio]{prederrorlevel0lF50L10M1_10.jpg}
%2461,width=2.67in,height=3.74in,keepaspectratio]{prederrorlevel0lF50L5_10M100.jpg}
%2461,width=2.67in,height=3.74in,keepaspectratio]{prederrorlevel1lF50L5_10M100.jpg}
  \includegraphics[width=2.67in,height=3.74in,keepaspectratio]{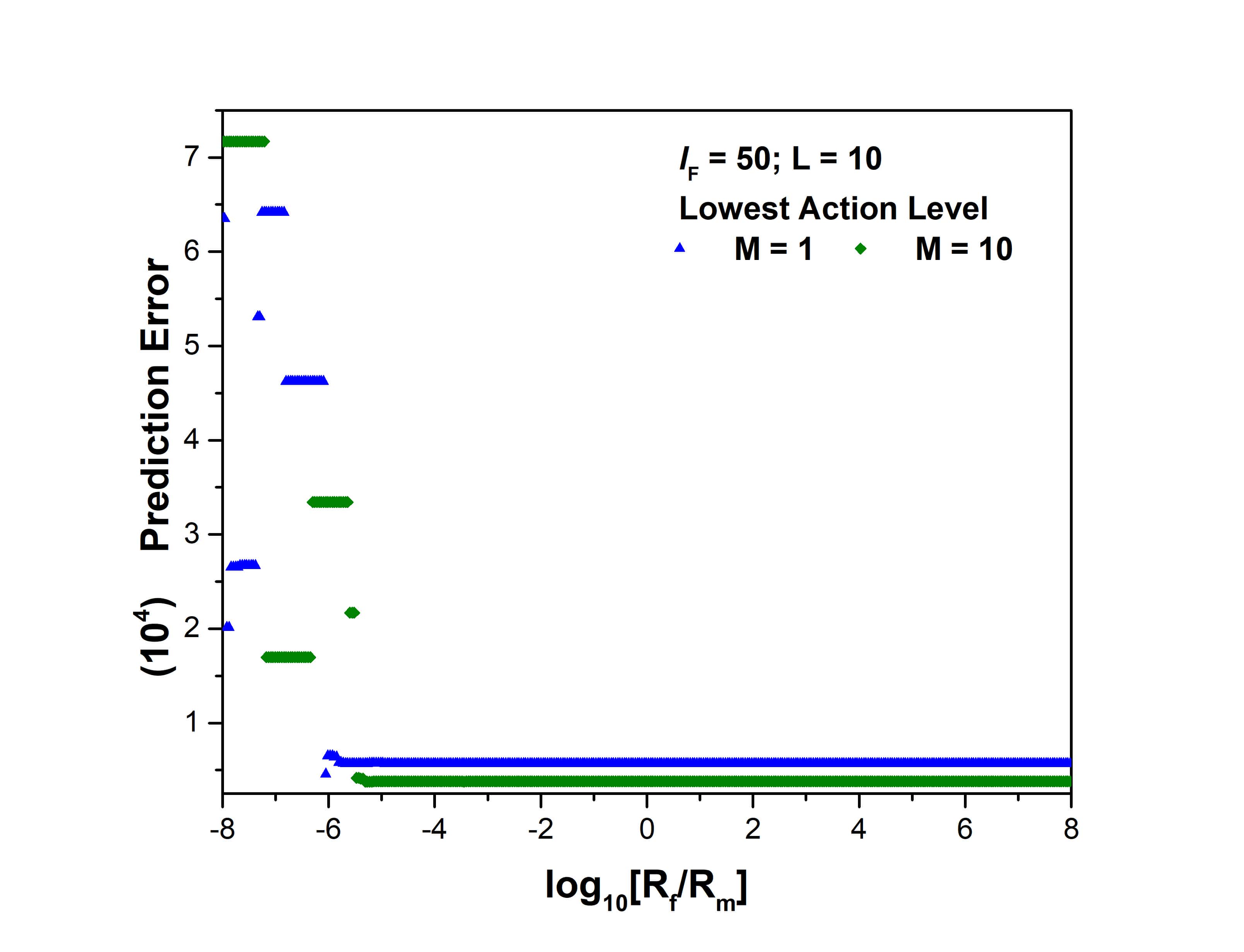}
  \includegraphics[width=2.67in,height=3.74in,keepaspectratio]{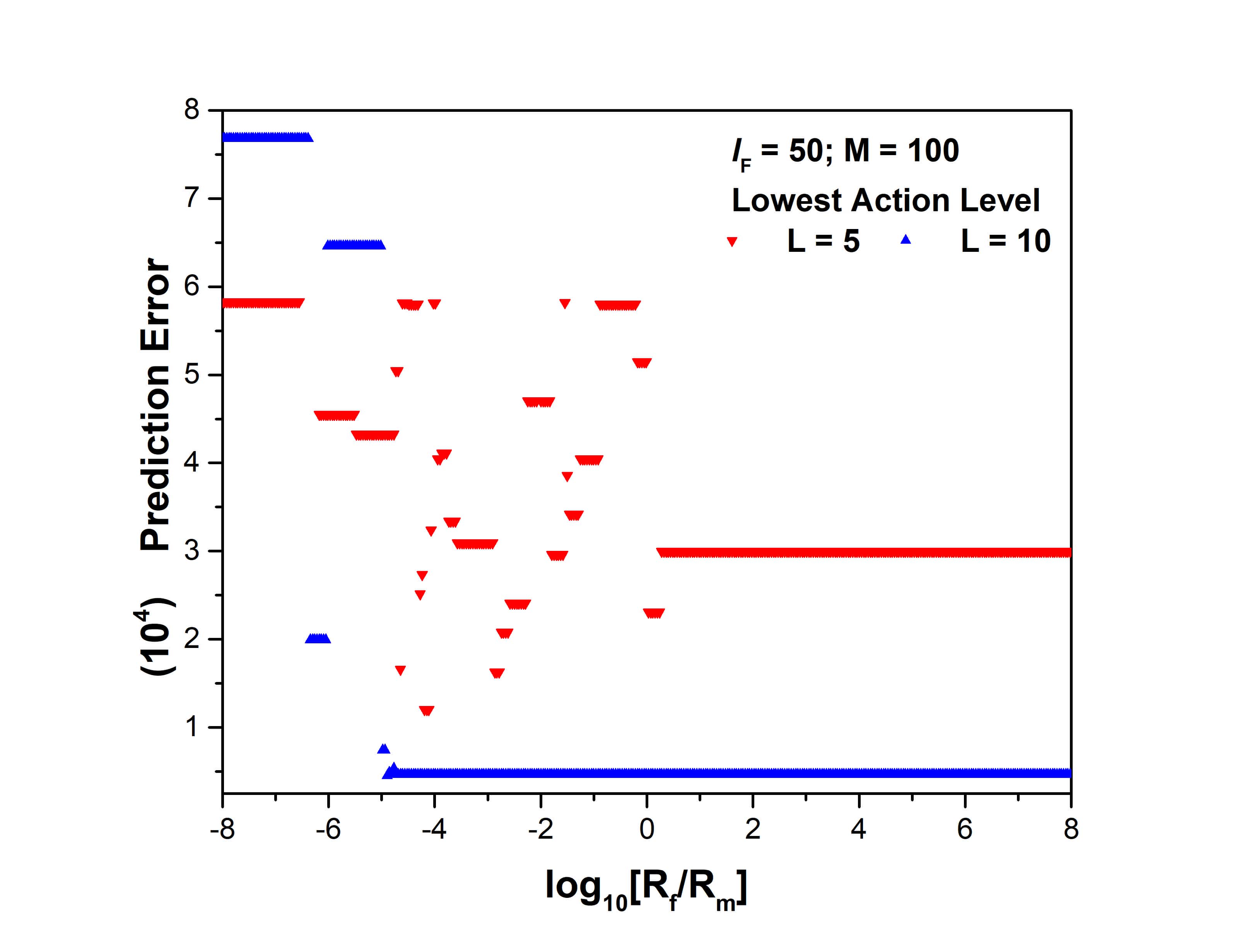}
  \includegraphics[width=2.67in,height=3.74in,keepaspectratio]{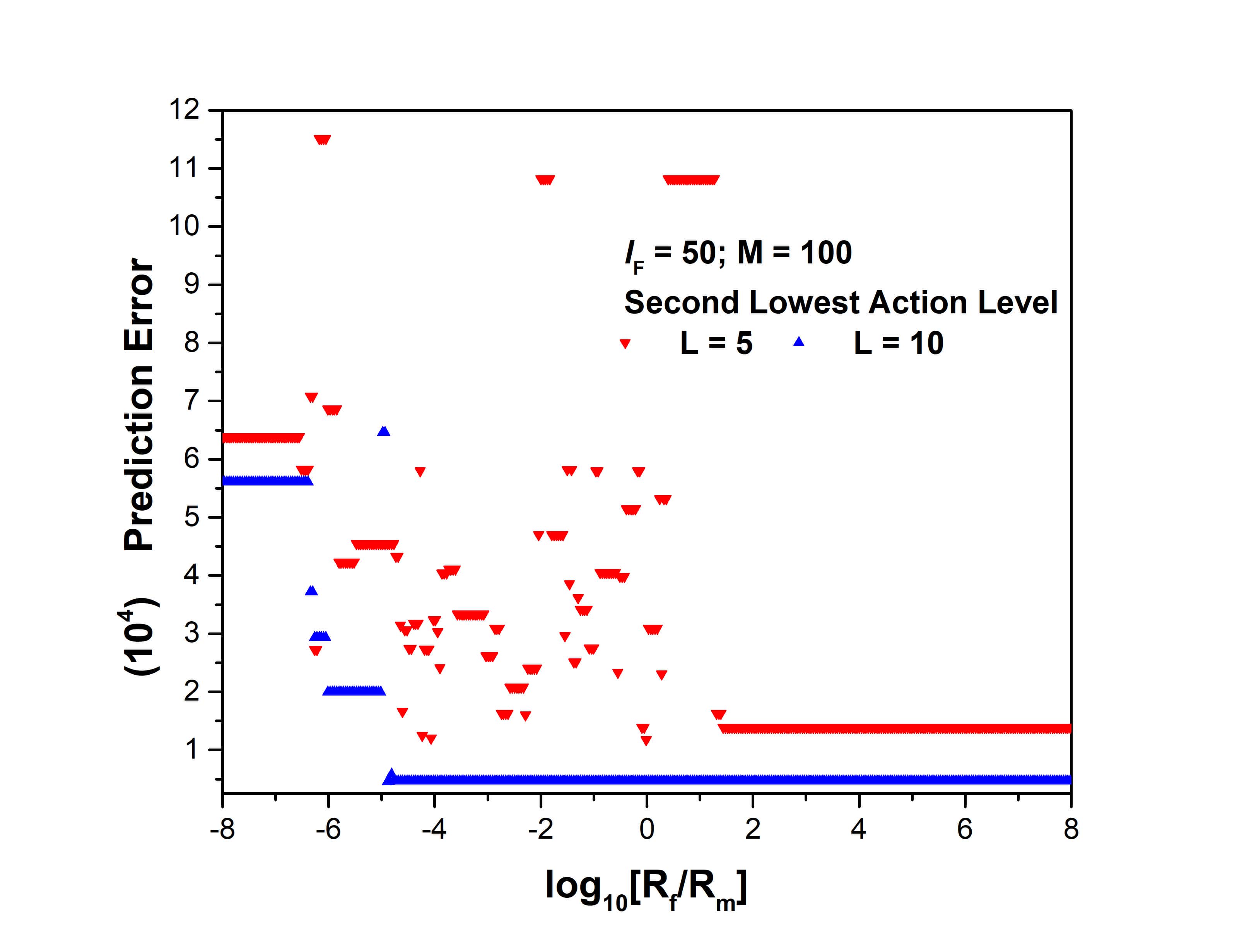}
  \caption{Prediction Errors for the AI/Machine Learning Network $\frac{1}{LM_P}\sum_{k=1}^{M_P} \sum_{r=1}^L (x^{(k)}_r(l_F) - y^{(k)}_r(l_F))^2$ averaged over $M_P$ new input/output pairs. A noisy input $y^{(k)}_r(l_0)$ produces the output $x^{(k)}_r(l_F)$ using the model estimated using the M training sets. This is compared with the output $y^{(k)}_r(l_F)$ produced with the original model used to produce the `data' for the twin experiment. In each case the number of neurons is N = 10 and $l_F = 50$. {\bf Top Left Panel} Using the model associated with lowest Action Level: L = 10 and M = 1 and M = 10. {\bf Top Right Panel} Using the model associated with lowest Action Level: L = 5 and L = 10; M = 100. {\bf Bottom Panel} Using the model associated with the second lowest Action Level: L = 5 and 10; M = 100.}
  \label{prederrorlevel0lF50L10M1_10}
\end{figure}

In Fig. (\ref{prederrorlevel0lF50L10M1_10}) we display the error in prediction after the mode, with $l_F = 50$ layers, has been set by the estimation of the weights. This error is constructed by selecting $M_P$ new input output pairs. Using each of the input elements for $L \le N$ components, we use the model with our estimated weights to evaluate $x^{(k)}_r(l_F)$ and compare that with $y^{(k)}_r(l_F)$ from each of the $M_P$ pairs. The square error averaged over $L$ presented components and over $M_P$ pairs 
\be 
\frac{1}{LM_P}\sum_{k=1}^{M_P} \sum_{r=1}^L (x^{(k)}_r(l_F) - y^{(k)}_r(l_F))^2
\ee 
is displayed. 

We see that increasing the information presented via increasing $L$ or $M$ leads to decreased average prediction errors when choosing the path corresponding to the lowest action level, {\bf Top Layers}, or choosing the path associated with the second lowest action level {\bf Bottom Panel}. The differences in quality of prediction (or generalization) in these examples among the cases analyzed is not large, and this has been noted~\cite{goodfellow16}.

%fig 3

%fig 4

%fig 5

%fig 6

\subsection{Recurrent Networks}

In this network architecture one allows both interactions among neurons from one layer to another layer as well as interactions among neurons within a single layer~\cite{jordan,elman}. 
The activity $x_j(l)$ of neuron j, j = 1,2,...,N in layer l $\{l_0,l_1, ..., l_F\}$ is given by $x_j(l) = f[\sum_i w_{ji}(l)x_i(l-1)]$ in a feedforward, {\tt layer goes to the next layer}, network. 

We can add interactions with a layer in the same fashion, and to give some `dynamics' to this within-layer activity we introduce a sequential label $\sigma$ to the activity of neuron j in layer l: $x_j(l,\sigma)$. The mapping from layer to layer and within a layer can be summarized by
\be 
x_j(l,\sigma) =  f[\sum_i w_{ji}(l)x_i(l-1,\sigma) + \sum_i W_{ji}(l)x_j(l,\sigma-1) ],
\ee
Another version of this allows the nonlinear function to be different for layer-to-layer connections and within-layer connections, so
\be 
x_j(l,\sigma) =  f[\sum_i w_{ji}(l)x_i(l-1,\sigma)]  + g[\sum_i W_{ji}(l)x_j(l,\sigma-1)],
\ee 
where $f(x)$ and $g(x)$ can be different nonlinear functions.

We can translate these expressions into the DA structure by recognizing that $x_j(l)$ is the `model variable' in the layer-to-layer function while in the recurrent network, the state variables become $x_j(l,\sigma)$. It seems natural that as dimensions of connectivity are added--here going from solely feedforward to that plus within-layer connections--that additional independent variables would be aspects of the `neuron' state variables' representation. 

In adding connections among the neurons within a layer we have another independent variable, we called it $\sigma$, and the `point' neurons depending on layer alone become {\tt fields} $x_j(l,\sigma)$. In the machine learning/AI networks we have no restrictions on the number of independent variables. This may lead to the investigation of `neural' fields $\phi_j(\v)$ where $\v$ is a collection of independent variables indicating which layers are involved in the progression of the field from an input to an output layer.

However many independent variables and however many `neurons' we utilize in the architecture of our model network, the overall goal of identifying the conditional probability distribution $P(\X|\Y)$ and estimating the moments or expected values of interest still comes down to one form or another in the approximation of integrals such as Eq. (\ref{expected}).

\subsection{Making Time Continuous; Continuous Layers: Deepest Learning}

There is much to learn about the data assimilation or machine learning problem as the number of layers or equivalently the number of time points within an epoch becomes very large. The limit of the action where the number of layers in an epoch becomes a continuous variable is, in data assimilation notation~\cite{qjrms17},
\small 
\bea 
&&A_0(\x(t),\dot{\x}(t)) = \int_{t_0}^{t_F} dt\, L(\x(t),\dot{\x}(t),t) \nonumber \\
&&L(\x(t),\dot{\x}(t),t) = \sum_{r=1}^L \frac{R_m(l,t)}{2}\biggl( x_r(t) - y_r(t) \biggr)^2 \nonumber
+ \sum_{a=1}^D \frac{R_f(a)}{2} \biggl ( \dot{x_a}(t) - F_a(\x(t))\biggr )^2.
\eea
\normalsize 
In this formulation the quantity $R_m(l,t)$ is non-zero only near the times $t \approx \tau_k$. It can be taken as proportional to $\delta(t-\tau_k)$.

Within the machine learning context we call this `deepest learning' as the number of layers goes to infinity in a useful manner.

The minimization of the action now requires that the paths $\x(t)$ in $\{\x(t),\dot{\x}(t)\}$ space satisfy the Euler-Lagrange equation
$\frac{d \blank}{dt}\biggl[\frac{\partial L(\x(t),\dot{\x}(t), t)}{\partial \dot{x}_a(t)}\biggr] = \frac{\partial L(\x(t),\dot{\x}(t), t)}{\partial x_a(t)}$, 
along with the boundary conditions $\delta x_a(t_0) p_a(t_0) = 0;\;\; \delta x_a(t_F) p_a(t_F) = 0$ 
where $ p_a(t) = \partial L(\x(t),\dot{\x}(t), t)/\partial \dot{x}_a(t)$ is the canonical momentum.

For the standard model, the Euler-Lagrange equations take the form
\be 
R_f[\frac{ d \blank}{dt}\delta_{ab} + DF_{ab}(\x(t))][\frac{dx_b(t)}{dt} - F_b(\x(t))] = R_m(l,t)\delta_{ar}(x_r(t) - y_r(t)),
\label{euler}
\ee
where we have $\DF(\x) = \partial \F(\x)/\partial \x$.

The E-L equations are the necessary condition, along with the accompanying boundary conditions, that show how errors represented on the right hand side of the E-L equation drive the model variables at all layers to produce $\x(l) \to \y(l)$ where data is available.

In the integral for $<G(\X)>$, the coordinates $\x(t_0)$ and $\x(t_F)$ are not restricted, so we have the `natural' boundary conditions~\cite{gfomin,kot,liberzon} $p_a(t_0) = 0$ and $p_a(t_F) = 0$.

This shows quite clearly that the minimization problem requires a solution of a two point boundary value problem in $\{\x(t), \v(t) = \dot{\x}(t)\}$ space. One way to address two point boundary value problems is to start at one end, $t_0$ with a value of $\x(t_0)$ and proceed from $t_F$ with a value of $\x(t_F)$ and integrate both ways requiring a match~\cite{press}. Furthermore, the residual of the measurement error term on the right hand side of Eq. (\ref{euler}) nudges the solution in $\x(t)$ to the desired output.

If one were to specify $\x(t_0)$, but not $\x(t_F)$, then the boundary conditions for the Euler-Lagrange equation are the given $\x(t_0)$ ($\delta \x(t_0) = 0$)
and require the canonical momentum $p_a(t_F) = 0$. Examining the Hamiltonian dynamics for this problem then suggest integrating the $\x(t)$ equation forward from $t_0$ and the canonical momentum equation backward from $t_F$. This is back propagation.

\subsubsection{Hamiltonian Dynamics Realization}

If one moves from the Lagrangian realization of the variational problem to a Hamiltonian version by trading in the phase space from $\{\x(t),\v(t)\}$ to
canonical coordinates $\{\x(t),\p(t)\}$, then the Hamiltonian $H(\x,\p)$ for the standard model reads
\small 
\be
H(\x,\p,t) = \sum_{a=1}^D\,\biggl\{ \frac{p_a(t)p_a(t)}{2 R_f(a)} + p_a(t)F_a(\x(t))\biggr\} -  \sum_{r=1}^L \, \frac{R_m(r,t)}{2}(x_r(t)-y_r(t))^2.
\ee
\normalsize 

In these coordinates the equations of motion are then given by Hamilton's equations
\small 
\bea 
&&\frac{dp_a(t)}{dt} = -p_b(t)\frac{\partial F_b(\x(t))}{\partial x_a(t)} + \delta_{ar}R_m(r,t)(x_r(t) - y_r(t)) \nonumber \\
&&\frac{dx_a(t)}{dt} = F_a(\x(t)) + \frac{p_a(t)}{R_f(a)}.
\label{hamgea}
\eea
\normalsize

Returning from this to discrete time (or layers) we see that if the variational principle is carried out in $\{\x,\p\}$ space, the boundary conditions $p_a(t_0) = p_a(t_F) = 0$ are quite easy to impose while the other variables, all the $x_a(t_k)$ and the $p_a(t_k);\;k \ne 0, F$, are varied. Going forward in $\x$ and backward in $\p$ is neither required nor suggested by this formulation. It is worth noting that in either $\{\x,\v\}$ space or $\{\x,\p\}$ space, the continuous time (layer) formulation has a symplectic symmetry~\cite{gfomin,qjrms17}. This not automatically maintained when the discrete time (layer) problem is reinstated~\cite{marsdenwest,marsden1997}; however, many choices of integration procedure in which time/layer becomes discrete and the symplectic symmetry is maintained are known~\cite{marsdenwest,marsden1997,hairer06}

In a detailed analysis~\cite{qjrms17,ye2015physrev} of the variational problem in Lagrangian and Hamiltonian formulations, it appears that the direct Lagrangian version in which the state variables $\x(t_n)$ or $\x(l_n)$ are varied, the symplectic structure can be maintained and the boundary conditions on the canonical momentum respected~\cite{marsdenwest,marsden1997}. 

In practice, this means that the direct variational methods suggested for the machine learning problems taking into account model error ($R_f \ne \infty$) may skirt issues associated with back propagation. 
This issue may be seen a bit more directly by comparing how one moves in $\{\x(t),\dot{\x}(t)\}$ space organized by Eq. (\ref{euler}) with the motion in $\{\x(t),\p(t)\}$ space guided by Eq. (\ref{hamgea}). These are equivalent motions of the model in time/layer, connected by a Legendre transformation from $\{\x(t),\dot{\x}(t)\} \to \{\x(t),\p(t)\}$. 

In the Hamiltonian form, where $R_f \to \infty$ is the limit where one usually works, moving in regions where $\DF(\x)$ may have saddle points may `slow down' the progression in the canonical momentum $\p(t)$. This may occur at a maximum, at a minimum, or at a saddle point of $\DF(\x)$. At any of these the observation in~\cite{deep15}: ``The analysis seems to show that saddle points with only a few downward curving directions are present in very large numbers, but almost all of them have very similar values of the objective function. Hence, it does not much matter which of these saddle points the algorithm gets stuck at." may apply. In the Lagrangian formulation Eq. (\ref{euler}) the manner in which $\DF(\x)$ enters the motion is quite different and may well avoid this confounding property. We have pointed out that Eq. (\ref{hamgea}) is backprop. The use of the Lagrangian variational principle~\cite{marsdenwest,marsden1997} solves the same problem, so may have an unexpected virtue.

\section{Summary and Discussion}

This paper has been directed to drawing a direct analogy between the formulation of a much utilized class of machine learning problems and a set of equivalent problems in data assimilation as encountered in many physical, biological and geoscience problems as well as in many engineering analyses where data driven model development and testing is a goal. The fundamental equivalence of the two inquiries is the core of this paper. 

The analogy allows us to identify methods developed in data assimilation as potentially quite useful in machine learning contexts. In particular the possibility of using variational annealing to produce the global minimum of the action (cost function) of the standard model of data assimilation with both observation error and model error appears potentially of value. 

The idea of making time continuous for purposes of exploring properties of data assimilation suggests a similar tactic in machine learning. The machine learning step of making layers continuous we have called ``deepest learning'' as deep learning appears to result from increasing the number of layers. In the continuous layer (time) formulation, we see clearly that the problem to be solved is a two point boundary value problem. This may lead to the construction and solution of tractable models that may helpfully illuminate how deep learning networks operate successfully and expand the possibilities of utilizing them employing additional methods for numerical calculations and for interpretation.

In the formulation of the statistical data assimilation problem at the general level expressed in Eq. (\ref{genaction}) we see that the measurement error term which is where information from data is passed to the model, it is explicitly information through the conditional mutual information that is being passed from observations to the model. This suggests that the idea that deep learning works because of major increases in computing power as well as in having large data sets, however, the attribute of the data sets is not so much as they are large but that they possess information, in a precise manner, that can be utilized to learn about models. The conjunction of information transfer and state and parameter estimation is embodied in the work of Rissanen~\cite{rissanen1989,rissanen2007} where he identifies the cost of estimating a parameter or a state variable at some time. The arguments we have presented suggest evaluating how much information in a data set is available to inform a model is of greater utility than just the size of the data set itself.

One point not made explicit in the main text, but worth noting, is that once we have formulated the data assimilation or machine learning problems as accurately performing high dimensional integrals such as Eq. (\ref{expected}), the Laplace approximation method, namely the usual variational principle, permits the investigation of corrections through further terms in the expansion of the action about the path leading to the global minimum. In~\cite{ye2015physrev} it is shown that corrections to this first approximation are small as $R_f$ becomes large when analyzing the standard model. This need not be the case for other choices of noise distributions in the measurement error or model error terms in the action.

Another item of interest is the argument noted in~\cite{deep15} that as the dimension of a model increases, one may find fewer and fewer local minima confounding the search in path space for a global minimum, and in that situation many more unstable saddle points in path space will 
arise~\cite{dauphin2014,aistats2015}. 

In the case of a chaotic system such as the Lorenz96 model, the evidence is that however large the dimension of the model itself and the paths over which one may search, there are multiple local minima until the number of measurements at any observation time is large enough and the information transferred to the model is sufficient. The role of the number of model evaluations between observations, suggested in some of the arguments here, also play a significant part in establishing whether the action surface has many local minima.

The view of a deep network as moving from a few hidden layers to many may also be illuminated by our arguments. One idea is that by increasing the number of hidden layers one is increasing the resolution in the analog of `time' in data assimilation. When one does that in data assimilation, we see it as probing the variation of the underlying model as it evolves from an initial condition through `layer' = `time.' Missing the higher frequency variations in time by employing a coarse grid in discrete time should have its counterpart role in the feedforward networks discussed here.

It is recognized that the `neuron' models widely utilized in machine learning applications have little in common with properties of biological neurons, the construction and implementation of large networks that have successful function within machine learning may prove a useful guide for the construction and implementation of functional natural neural networks.

Finally, it is important to comment that while the analogy drawn and utilized here may improve the testing and validation of models supported by observational data, it does not assist in the selection of the models and their formulation. That is still a task to be addressed by the user.

\section*{Acknowledgments} We express our appreciation to our colleagues Dan Breen and Jeff Elman for important discussions. Also the CENI team made finding the analogs in this paper possible. Many thanks to Tim Gentner, Gert Cauwenberghs, and especially Gabe Silva improved the manuscript. Jon Shlens provided a detailed view from the active world of machine learning. Partial support from the MURI Program (N00014-13-1-0205) sponsored by the Office of Naval Research is acknowledged as is support for A. Shirman from the ARCS Foundation.

\newpage 

%\bibliographystyle{plain}
%\bibliographystyle{alpha}
%\bibliographystyle{apalike}
%\bibliography{symplecanneal0901}

\end{document}